\documentclass[runningheads]{llncs}

\usepackage{eccv}

\usepackage{eccvabbrv}

\usepackage{graphicx}
\usepackage{booktabs}

\usepackage[accsupp]{axessibility}  

\usepackage{hyperref}

\usepackage{orcidlink}
\usepackage{xcolor}         
\usepackage{bm}
\usepackage{multirow}\usepackage{makecell}
\usepackage{array}
\usepackage{cite}
\usepackage{graphicx}

\makeatletter
\def\thanks#1{\protected@xdef\@thanks{\@thanks
        \protect\footnotetext{#1}}}
\makeatother
\usepackage[marginal]{footmisc}
\usepackage[misc]{ifsym}

\usepackage{caption}
\title{Rethinking Weakly-supervised Video Temporal Grounding From a Game Perspective}

\author{\small
Xiang Fang\inst{1}\textsuperscript{*} \and Zeyu Xiong\inst{1}\textsuperscript{*} \and Wanlong Fang\inst{1}\textsuperscript{*} \and Xiaoye Qu\inst{1} \and Chen Chen\inst{2} \and Jianfeng Dong\inst{3} \and Keke Tang\inst{4} \and Pan Zhou\inst{1}\textsuperscript{\Letter} \and Yu Cheng\inst{5} \and Daizong Liu\inst{6}\textsuperscript{\Letter}
\thanks{
\textsuperscript{*} Xiang Fang, Zeyu Xiong, and Wanlong Fang contributed equally to this work. \\
\textsuperscript{\Letter} Corresponding authors: Pan Zhou and Daizong Liu.}
}
\institute{\small
Hubei Key Laboratory of Distributed System Security, Hubei Engineering Research Center on Big Data Security, School of Cyber Science and Engineering, Huazhong University of Science and Technology \and  University of Central  Florida  \and  Zhejiang Gongshang University \and  Guangzhou University \and  The Chinese University of Hong Kong \and Peking University\\
\email{xfang9508@gmail.com, zeyuxiong@hust.edu.cn, wanlongfang@gmail.com, xiaoye@hust.edu.cn, chen.chen@crcv.ucf.edu, dongjf24@gmail.com, tangbohutbh@gmail.com, panzhou@hust.edu.cn, chengyu@cse.cuhk.edu.hk, dzliu@stu.pku.edu.cn}}

\titlerunning{Rethinking WS-VTG From a Game Perspective}
\authorrunning{X. Fang, Z. Xiong, W. Fang et al.}

\begin{document}
\captionsetup{font={small}}
\maketitle

\begin{abstract}
This paper addresses the challenging task of weakly-supervised video temporal grounding.
Existing approaches are generally based on the moment proposal selection framework that utilizes contrastive learning and reconstruction paradigm for scoring the pre-defined moment proposals.
Although they have achieved significant progress, we argue that their current frameworks have overlooked two indispensable issues:
1) Coarse-grained cross-modal learning: previous methods solely capture the global video-level alignment with the query, failing to model the detailed consistency between video frames and query words for accurately grounding the moment boundaries.
2) Complex moment proposals: 
their performance severely relies on the quality of proposals, which are also time-consuming and complicated for selection.
To this end, in this paper, we make the first attempt to tackle this task from a novel game perspective, which effectively learns the uncertain relationship between each vision-language pair with diverse granularity and flexible combination for multi-level cross-modal interaction.
Specifically,  we creatively model each video frame and query word as game players with multivariate cooperative game theory to learn their contribution to the cross-modal similarity score.
By quantifying the trend of frame-word cooperation within a coalition via the game-theoretic interaction, we are able to value all uncertain but possible correspondence between frames and words.
Finally, instead of using moment proposals, we utilize the learned query-guided frame-wise scores for better moment localization.
Experiments show that our method achieves superior performance on both Charades-STA and ActivityNet Caption datasets.
\end{abstract}

\vspace{-6mm}
\section{Introduction}
Video temporal grounding (VTG) aims to localize the start and end timestamps of a specific event moment described by a given language query in an untrimmed video \cite{anne2017localizing,gao2017tall,wang2022learning,wangchangshuo20223d,wang20233d,wang2021brief,jiang2023lttpoint,zhang2024pointgt,zhang2023deep,ning2023occluded,ning2024enhancement,ning2023pedestrian,yu2024pedestrian}. This task enables us to efficiently find video moments of human interest instead of traversing the entire video, which has broad application potential in video surveillance \cite{dong2022partially,liu2021context,zheng2023progressive,zheng2022cpl,dong2022reading,dong2018predicting,dong2022dual,liu2023incomplete,liu2023dicnet,liu2023information,tang2022decision,tang2024effective,tang2022rethinking,tang2023reppvconv}, video summarization \cite{guo2024benchmarking,dong2023region,dong2023hierarchical,lin2023univtg,10447574,2023A,10086538,10042189,fang2020double,liu2023exploring,wang2025taylor,fang2026towardsicml,kuai2026dynamic,wang2025point,fang2025your,zhang2025monoattack,fang2023hierarchical,liu2024towards,yang2025eood,fang2022multi,fang2026cogniVerse,lei2025exploring,fang2023you,wang2025dypolyseg,fang2025hierarchical,yan2026fit,fang2025adaptive,wang2026topadapter,cai2025imperceptible,fang2026slap,wang2026reasoning,fang2026immuno,wang2026biologically,fang2026disentangling,wang2025reducing,fang2026advancing,fang2026unveiling,wang2026from,liu2023conditional,liu2026attacking,fang2026rethinking,wang2025seeing,fang2026towards,fang2025multi,fang2024fewer,liu2024pandora,fang2024multi,fang2025turing,fang2024not,liu2023hypotheses,liu2024unsupervised,fang2023annotations,xiong2024rethinking,fang2021unbalanced,wang2025prototype,zhang2025manipulating,fang2026align,tang2024reparameterization,fang2025adaptivetai,tang2025simplification,fang2021animc,cai2026towards,fang2020v}, \textit{etc}. Most VTG methods \cite{liu2024towards,liu2024unsupervised,liu2024transform,fang2023annotations,liu2023conditional,liu2023filling,liu2023exploring,liu2023hypotheses,fang2023you,liu2022few,fang2022multi,fang2023hierarchical,zhu2023rethinking,liu2022reducing,liu2022skimming} follow the fully-supervised setting, where each frame is manually annotated as query-relevant or not.
Although these methods achieve significant breakthroughs, they severely rely on extensive laborious manual annotations of moment boundaries, thus limiting their scalability and practicability in real-world applications. 
To alleviate the dense reliance,  weakly-supervised VTG (WS-VTG) has received increasing attention \cite{mithun2019weakly,chen2020look,song2020weakly,tan2021logan,lin2020weakly,zhang2020counterfactual}, which only requires the knowledge of matched video-query pairs without detailed frame-wise annotations, which is more challenging than the fully-supervised setting.

\begin{figure}[t]   
	\centering
	\includegraphics[width=\linewidth]{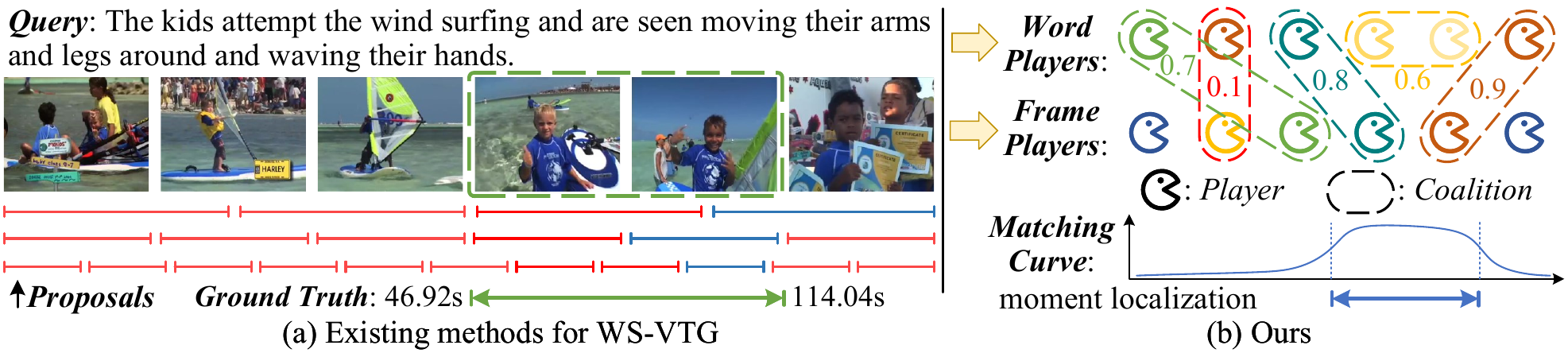}
	\vspace{-7mm}
	\caption{  \small{(a) Most existing WS-VTG methods define moment proposals for query matching. They not only suffer from the coarse-grained alignment but also are limited to the quality of the proposals. (b) Instead of using proposals, our method reviews the WS-VTG from a brand-new game-interaction perspective to learn more fine-grained frame-aware self- and cross-modal alignment for accurate boundary localization.}
	} 
	\label{fig:intro}
	\vspace{-6mm}
\end{figure}

Since there is no frame-wise annotation in WS-VTG, learning detailed frame-query alignment is difficult. 
As shown in \cref{fig:intro}(a), almost all existing WS-VTG solutions directly employ the proposal-based framework \cite{mithun2019weakly,gao2019wslln,chen2020look,tan2021logan} that first generates multiple moment proposals and then learns the scores to indicate the potential alignment between video proposals and language query.
Some approaches further utilize reconstruction-based paradigm \cite{song2020weakly,lin2020weakly,zhang2020counterfactual} to minimize the reconstruction loss between the partially masked query and moment proposals to identify the proposal which best reconstructs the query.
Although exising WS-VTG methods have achieved great progress, we argue that they still suffer from two inescapable limitations:
1) Firstly, 
as frame-wise annotation is not provided, 
these WS-VTG methods generally generate the moment proposals for all videos via sliding windows. In this way, they only
model the coarse-grained cross-modal interaction of each proposal-query pair.
Actually, in most cases, we expect to capture more fine-grained information for accurate boundary localization, such as how much the semantics of a specific frame helps or harms the query-guided cross-modal alignment. 
Unfortunately, existing methods rely on proposal-level cross-modal learning cannot achieve this goal.
2) {Secondly, the performance of these proposal-based methods is
severely limited by the quality of moment proposals as they devise moment proposals regardless of the specific contents and difficulty
of each video. 
}


To tackle the above limitations, in this paper, we propose to address the WS-VTG task from a new game-theory perspective, which creatively models the subtle video frames and query words as game players to learn their uncertain but possible cross-modal interaction with diverse granularity and flexible combination.
The game learning helps to generate query-guided frame-aware knowledge for fine-grained video grounding.
As shown in \cref{fig:intro}(b), if visual representations and textual representations have strong semantic correspondence, they tend to cooperate together and contribute to a high cross-modal similarity score.
By forming these cooperated frame-word representations as a coalition and quantifying the trend of cooperation within a coalition via the game-theoretic interaction index, we can not only learn the coalition contribution to the cross-modal semantic similarity, but also measure the additional benefits brought by the coalition compared with the costs of the lost coalitions of these players with others. In this manner, our cooperative WS-VTG game is able to value all possible correspondence between frames and words for sensitive and explainable cross-modal contrast. 
By formulating these detailed cross-modal correspondences as fine-grained query-guided frame-wise scores, we can efficiently generate the moment instead of using complicated proposals.
In particular, we implement the cooperative game in both self-modal and cross-modal scenarios for self-modal attentive learning and cross-modal fine-grained alignment. We further extend the cross-modal game with multi-level query-semantic interaction for more comprehensive video understanding.

Our main contributions are summarized as follows:
\vspace{-2pt}
\begin{itemize}
    \item 
    We reveal the limitations of existing WS-VTG methods, which rely on complex moment proposals and fail to capture fine-grained frame-word alignment for accurate boundary grounding. To this end, 
    we propose a novel game-theory based framework to learn the detailed frame-wise query-relevance scores for efficiently constructing the accurate moment. 
    \item We apply the game learning to both self- and cross-modal scenarios for contextual self-enhancement and cross-alignment. The cross-modal game is also extended to multi-level for comprehensive learning.
    \item To verify the effectiveness of our framework, we conduct experiments on two challenging WS-VTG benchmarks, \textit{i.e.}, Charades-STA and ActivityNet Caption. Experiments show that our method achieves superior performance. In representative cases, our method outperforms all compared methods by 4.53\% on Charades-STA.
\end{itemize}

\section{Related Works}
\label{sec:relate}

\noindent \textbf{Fully-supervised video temporal grounding.}
Existing methods generally address the video temporal grounding in a fully-supervised manner, where both the annotations of video-sentence pairs and corresponding moment boundaries are given.
Most of them \cite{gao2017tall,Alpher07,wang2020temporally,yuan2020semantic,zhang2019man,zhang2019cross,Alpher33,liu2023jointly,xiong2023tracking,xiong2022gaussian,liu2022learning,guo2022hybird,liu2023conditional,liu2024towards,tang2024reparameterization} utilize the proposal-based framework,
which first integrate the sentence representation with those pre-defined moment proposals individually, and then evaluate their matching relationships. 
The proposal with the highest matching score is selected as the prediction. 
Instead of utilizing the moment proposals,
recent proposal-free methods \cite{Alpher08,mun2020local,zeng2020dense,chen2020rethinking} directly regress the temporal locations of the target moment. 

\noindent \textbf{Weakly-supervised video temporal grounding.}
As manually annotating temporal boundaries of target moments is time-consuming, recent research attentions shift to weakly-supervised video temporal grounding \cite{mithun2019weakly,chen2020look,song2020weakly,tan2021logan}, which only requires video-level annotations. Mithun \textit{et al.} \cite{mithun2019weakly} proposed the first weakly-supervised model to learn a joint embedding space for video and query representations. Gao \textit{et al.} \cite{gao2019wslln} developed a two-stream structure to measure the moment-query consistency and conduct moment selection simultaneously.
Although the above methods have achieved promising performance, they are two-stage approaches that utilize multi-scale sliding windows to generate moment candidates, therefore suffering from inferior effectiveness and efficiency.
To address this issue, \cite{lin2020weakly,zhang2020counterfactual,ma2020vlanet} score all the moments sampled at different scales in a single pass and further improve the moment-sentence matching accuracy. 
However, almost all of them rely on moment proposals for matching and selection, which fail to capture and distinguish more fine-grained details among visually similar frames for acquiring more accurate moment boundaries.


\noindent \textbf{Cooperative game theory.}
Cooperative game theory focuses on the formation of coalitions and cooperation among rational players or groups to achieve common goals~\cite{osborne1994course,chalkiadakis2011computational,torsello2006grouping,albarelli2010game,patel2021game,li2022fine,jin2021game,jin2023video,ma2017forecasting,pavan2003new}. A typical cooperative game consists of a set of players with a game function. The game function maps all possible subsets of players, called group or coalition, to a number which represents the total payoff earned by these players working cooperatively to achieve the goal \cite{rodola2012game,dowdall2006coalitional,li2023g2l,donoser2013diffusion}. To measure the contributions of each player and allocate different payoffs to these individuals fairly, a few researchers have proposed various concepts of value by computing the average added worth that one player brings to all possible coalitions, such as Shapley value~\cite{shapley1953value,winter2002shapley,michalak2013efficient} and Banzhaf value~\cite{banzhaf1964weighted,lehrer1988axiomatization,nowak1997axiomatization}. 
Recently, some concepts of cooperative game theory have been interpreted in the field of deep learning~\cite{lundberg2017unified, zhang2021interpreting, datta2016algorithmic}. 
For example, Li \textit{et al.}~\cite{li2022fine} propose a fine-grained image-text semantic alignment pre-training framework based on Shapley interaction. 
Different from these works, we propose for the first time to model the detailed consistency between video frames and query words by recasting multivariate game-theoretic indices to address more complicated WS-VTG task.

\begin{figure*}[t]
        
	\centering
	\includegraphics[width=\linewidth]{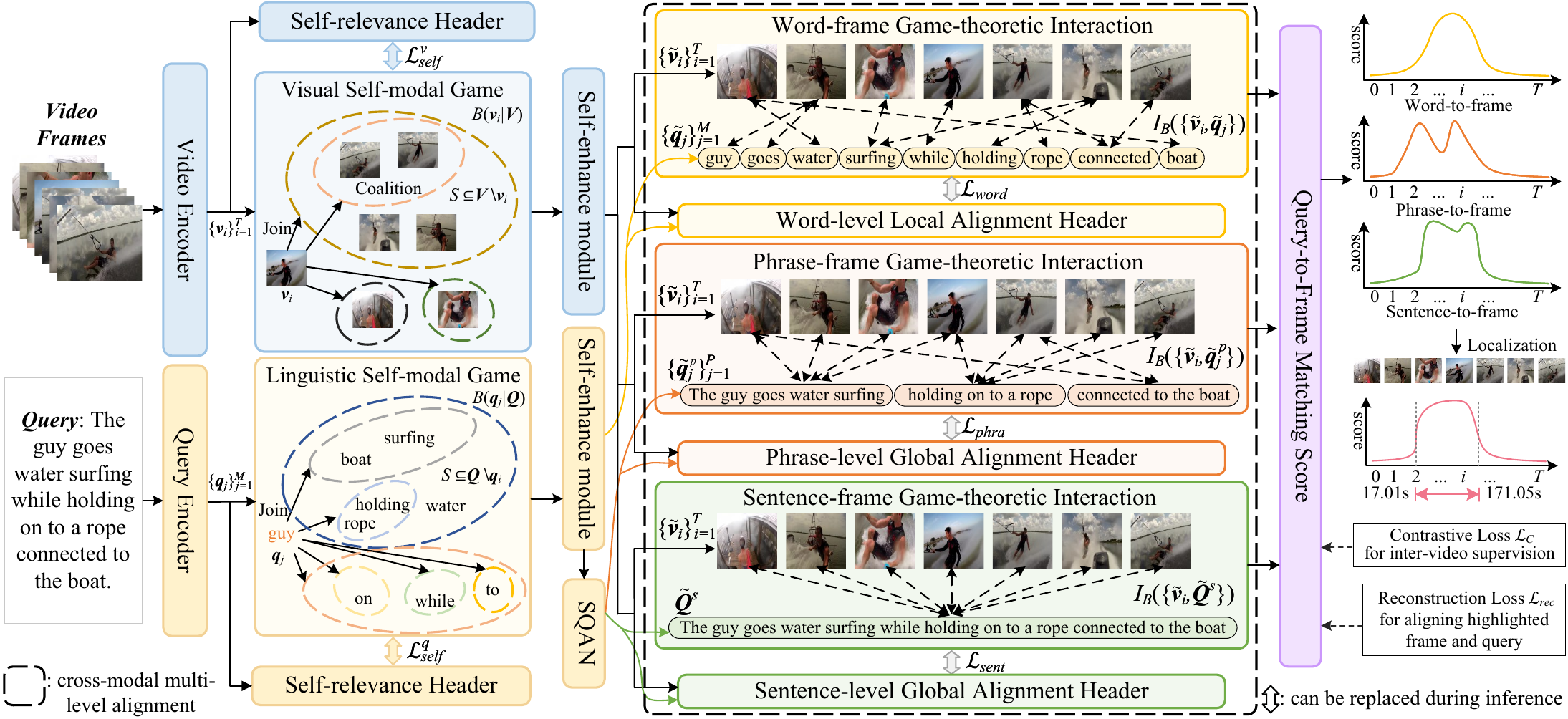}
	\vspace{-6mm}

	\caption{  \small{Overview of our framework, where ``SQAN'' denotes Sequential Query Attention Network. Specifically, we first conduct self-modal game to enhance the self-modal instance semantics, then devise multi-level cross-modal games to learn the fine-grained and uncertain cross-modal correspondence. At last, we predict the moment based on the query-guided frame-wise scores.}
	} 
	\label{fig:pipe}
	\vspace{-6mm}
\end{figure*}

\section{Methodology}
\label{headings}
In this section, we elaborate on the proposed method, 
which learns the uncertain and fine-grained cross-modal alignment in WS-VTG task. 
Specifically, our method conceptualizes the interaction between the video and query as a game, where each frame/word acts as a player and forms coalitions. 
During the game learning, visual frame and linguistic word representations with strong semantic relevance will cooperate to form a new coalition, the benefits of which are quantified through the game interaction index. Our method maximizes payoff by strategically forming coalitions to achieve fine-grained semantic alignment between video and query.
The overall framework is shown in \cref{fig:pipe}.


\subsection{Preliminaries}
\noindent \textbf{Problem formulation.}
Generally, given an untrimmed video and a sentence query, WS-VTG aims to localize the specific video moment semantically related to the query with only the video-level annotation. Let $V=\{v_1,...,v_i...,v_T\}$ denote the untrimmed video with $T$ frames, and $Q=\{q_1,...,q_j,...,q_M\}$ represent the sentence query containing $M$ words, where $v_i$ and $q_j$ are the $i$-th frame and $j$-th word, respectively. The objective function $\mathcal{F} $  can be formulated as $\mathcal{F}: (V, Q) \rightarrow (\tau_s, \tau_e)$, where $\tau_s,\tau_e$ denote the start and end timestamps of the video moment boundary in video $V$ semantically corresponding to  query $Q$. 

\noindent \textbf{Video and query representations.} 
To extract contextual video representation, following \cite{gao2017tall,zhang2020span}, each input video $V$ is fed into a pre-trained 3D convolutional network \cite{carreira2017quo,tran2015learning} with a stack of multi-head self-attention~\cite{vaswani2017attention} layers for video encoding, then we get frame-wise video representation  $\bm{V} =\{ \bm{v}_i\}_{i=1}^{T} \in \mathbb{R} ^{T\times D_h}$, where $D_h$ is the hidden dimension. For query representation, we also follow  \cite{zhang2020span,liu2021context} to embed each word of the query by the Glove~\cite{pennington2014glove} model with a further multi-head self-attention module and a BiLSTM layer. Similarly, we can obtain the word-wise query representation $\bm{Q} =\{ \bm{q}_j\}_{j=1}^{M} \in \mathbb{R} ^{M\times D_h}$.

\subsection{Game-theoretic Learning for Self-modal Semantic Enhancement}
We argue that the instance relations within each modality are  important to infer the consecutive visual event contents or linguistic phrase contexts. 
To achieve this self-modal learning, we review previous classic game-theoretic value \cite{shapley1953value,grabisch1999axiomatic} to explore the weighted semantic contribution of each frame/word to the whole video/query semantics for enhancing the self-modal representation of both video and query.

\noindent \textbf{Visual self-modal game.} Generally, a cooperative game consists of a player set $\mathcal{P}=\{1,2,...,n\}$ and a game function $g(\cdot)$. In detail,  function $g(\cdot)$ maps each player subset $\mathcal{S} \subseteq \mathcal{P}$ to a score value, which indicates the payoff when all players in coalition $\mathcal{S} $ work together in the game. To capture the inherent correlations within the video modality, we directly take the video frames $\bm{V} =\{ \bm{v}_i\}_{i=1}^{T}$ as frame players $\mathcal{P} = \bm{V},n=T$ to play the game and evaluate frame-wise contribution to frame-level video semantics. 

Since different frames contribute and weight differently to video understanding, we attempt to adopt a game-theoretic value to measure the importance of each frame player in the cooperative game. Specifically, each frame player $\bm{v}_i$ in player set $\bm{V}$ has a weighted marginal contribution to the entire game, which can be measured by the game-theoretic value (\textit{e.g.,} Banzhaf value~\cite{grabisch1999axiomatic}) as:
\small
\begin{equation}
B(\bm{v}_i|\bm{{V}}) = \frac{1}{2^{T-1}} \sum_{\mathcal {S} \subseteq  \bm{V} \backslash  \bm{v}_i}[g(\mathcal {S} \cup \{\bm{v}_i\})-g(\mathcal{S})],
\label{banzhafvalue}
\end{equation}\normalsize
where $\mathcal {S} \subseteq \bm{V} \backslash  \bm{v}_i$ represents a group formed without $\bm{v}_i$, $\frac{1}{2^{T-1}}$ is the likelihood of $\mathcal {S}$ being sampled, and $g(\mathcal {S} \cup \{\bm{v}_i\})-g(\mathcal{S})$ denotes the marginal contribution brought by player $\bm{v}_i$ joining the coalition $S$.
$B(\bm{v}_i|\bm{{V}})$ indicates the ability of player $\bm{v}_i$ to influence the video-level semantic by calculating the weighted marginal contribution of player $\bm{v}_i$ to all possible coalitions $\mathcal {S} $. 
Generally, within the context of our visual game, coalitions in the video modality represent consecutive frames of the same event or frames that are semantically similar. In this way, we can enumerate and learn all possible relations between the frames within the video for more contextual visual feature learning. 

\noindent \textbf{Visual soft supervision.}
To assist the above game-theoretic value learning,
we first develop an additional learnable video-domain relevance header to predict the same self-modal relevance $\hat{{R}}^v = \{ \hat{{r}}_i^v\}_{i=1}^{T} = header(\bm{V})$ as Banzhaf values $\{B(\bm{v}_i|\bm{{V}})\}_{i=1}^T$ among all frames for soft supervision (more discussions will be illustrated later). 
Specifically, the header is implemented by: 1) 1D convolutional layers for local self-consistency modeling, 2) a self-attention module for global self-relevance capturing, 3) a convolutional layer for decoding. Then, we optimize the Kullback-Leibler Divergence (KLD) between the predicted relevance $ \hat{{r}}_i^v$ and the Banzhaf value $B(\bm{v}_i|\bm{{V}})$ for learning the consistent self-modal similarity.
In detail, we define the game-guided frame-wise probability distribution as $D^{v}=[p_1^{v},p_2^{v},...,p_i^{v},...,p_T^{v}]^\top$, where $p_i^{v}= \frac{\exp{(B(\bm{v}_i|\bm{V}))}}{\sum_{l=1}^T \exp {(B(\bm{v}_l|\bm{{V}}))}}$. Similarly, the predicted frame-wise probability distribution calculated by the output of the self-relevance header can be denoted as $\hat{D}^{v}=[\hat{p}_1^{v},\hat{p}_2^{v},...,\hat{p}_i^{v},...,\hat{p}_T^{v}]^\top$, where $\hat{p}_i^{v}= \frac{\exp{(\hat{{r}}_i^v)}}{\sum_{l=1}^T \exp {(\hat{{r}}_l^v)}}$. Finally, the self-relevance loss $\mathcal{L}_{self}^v$ for each video is:
\small
\begin{equation}
\mathcal{L}_{self}^{v}={\rm KLD}(\hat{D}^{v} || D^{v})=-\frac{1}{T}\sum\nolimits_{i=1}^T p_i^{v}(\log \hat{p}_i^{v} - \log p_i^{v}).
\label{selfKLD}
\end{equation}\normalsize
Based on the learned game-theoretic value $B(\bm{v}_i|\bm{{V}})$ for each player $\bm{v}_i$ in player set $\bm{V}$, we
deploy a Softmax function to obtain the self-modal enhanced video features $\widetilde{\bm{v}}_i$ as:
\small
\begin{equation}
\widetilde{\bm{v}}_i =  \bm{v}_i + \sum_{\mathcal {S} \subseteq  \bm{V} \backslash  \bm{v}_i}{\rm Softmax}[g(\mathcal {S} \cup \{\bm{v}_i\})-g(\mathcal{S})] \cdot (\mathcal {S} \cup \{\bm{v}_i\}).
\label{enhancev}
\end{equation}\normalsize
The enhanced video features $\widetilde{\bm{V}} = \{\widetilde{\bm{v}}_i\}_{i=1}^T$ aggregates the relevant contexts from the whole video.

\noindent \textbf{Linguistic self-modal game.} Similarly, we deem query words $\bm{Q} =\{ \bm{q}_j\}_{j=1}^{M}$ as players $\mathcal{P} = \bm{Q},n=M$ and calculate each word's marginal contribution $B(\bm{q}_j|{\bm{Q}})$ to the whole query $\bm{Q}$ by formulating the linguistic self-modal game as:
\small
\begin{equation}
B(\bm{q}_j|{\bm{Q}}) = \frac{1}{2^{M-1}} \sum_{\mathcal {S} \subseteq  \bm{Q} \backslash  \bm{q}_j}[g(\mathcal {S} \cup \{\bm{q}_j\})-g(\mathcal{S})].
\label{banzhafvalueq}
\end{equation}\normalsize
\noindent \textbf{Linguistic soft supervision.}
We first take the game probability distributions $D^{q}=[p_1^{q},p_2^{q},...,p_j^{q},...,p_M^{q}]^\top$ as the soft label, and then supervise the predicted distributions $\hat{D}^{q}=[\hat{p}_1^{q},\hat{p}_2^{q},...,\hat{p}_j^{q},...,\hat{p}_M^{q}]^\top$ of another header by KLD loss $\mathcal{L}_{self}^q$ similar to \cref{selfKLD}.  
At last, we can obtain the enhanced query features $\widetilde{\bm{Q}} = \{\widetilde{\bm{q}}_j\}_{j=1}^{M}$ by: 
\small
\begin{equation}
\widetilde{\bm{q}}_j =  \bm{q}_j + \sum_{\mathcal {S} \subseteq  \bm{Q} \backslash  \bm{q}_j}{\rm Softmax}[g(\mathcal {S} \cup \{\bm{q}_j\})-g(\mathcal{S})] \cdot (\mathcal {S} \cup \{\bm{q}_j\}).
\label{enhanceq}
\end{equation}\normalsize

\noindent \textbf{Discussion on  learnable header.}
Although we can directly utilize the game interaction value to enumerate the relations between different players via \cref{banzhafvalue,banzhafvalueq}, there are still two challenges:
1) No supervision signal: The performance of the game interaction severely depends on the quality of the learned player features. However, there is no supervision for learning the representative features.
2) Complex and time-consuming: The game process is complex and costs much time and resources.
To this end, we develop a learnable header to mimic the game interaction by utilizing the KLD loss. This KLD function serves as a soft supervision label to not only learn the consistency between the game interaction and header, but also potentially train more distinguishing player features of each modality. During the inference, we can solely utilize the prediction of header to model the game interaction instead of \cref{banzhafvalue,banzhafvalueq}.

\subsection{Game-theoretic Interaction for Cross-modal Multi-level Alignment}
After obtaining the self-enhanced video and query features, we further develop a cross-modal game between frames and words to handle their uncertainty during their fine-grained semantic alignment.

\noindent \textbf{Cross-modal game.}
From a game-theoretic perspective, we attempt to conceptualize the problem of fine-grained cross-modal alignment as a collaborative effort between video frames and query words. Specifically, when a video frame and a query word exhibit a high degree of semantic similarity, they are more likely to collaborate and their union will have a larger score contributed to the final moment. To this end, we take the enhanced video features $\widetilde{\bm{V}} =\{ \widetilde{\bm{v}}_i\}_{i=1}^{T}$ and enhanced query features $\widetilde{\bm{Q}} =\{ \widetilde{\bm{q}}_j\}_{j=1}^{M}$ as two different kinds of players $\mathcal{P} =\{ \widetilde{\bm{v}}_i\}_{i=1}^{T} \cup \{ \widetilde{\bm{q}}_j\}_{j=1}^{M}, n=T+M$.
In a cooperative game, different players tend to work together in groups, called coalitions, to achieve a common goal, \textit{i.e.}, representing the same semantics of the target event. In the case of two players $\widetilde{\bm{v}}_i$ and $\widetilde{\bm{q}}_j$ in set $\mathcal{P}$, it may occur that $g (\{\widetilde{\bm{v}}_i\})$ and $g (\{\widetilde{\bm{q}}_j\})$ are small individually, but at the same time the reward of their forming coalition $g (\{\widetilde{\bm{v}}_i,\widetilde{\bm{q}}_j\})$ is considerable. This is because players in the coalition interact and collaborate with each other, which may bring additional benefits (or costs) to the game. 
Therefore, we also need to measure the additional benefits brought by the coalition compared with the costs of the lost coalitions of these players with others. 
To  form a better coalition $\{\widetilde{\bm{v}}_i,\widetilde{\bm{q}}_j\}\subseteq \mathcal{P}$,
we utilize the effective interaction index $I_B(\{\widetilde{\bm{v}}_i,\widetilde{\bm{q}}_j\})$ like~\cite{grabisch1999axiomatic} to compute the additional benefit as follows:
\small
\begin{align}
I_B(\{\widetilde{\bm{v}}_i,\widetilde{\bm{q}}_j\})\! = \!\frac{1}{2^{n-2}}\! \sum_{\mathcal {S} \subseteq  \mathcal{P} \backslash  \{\widetilde{\bm{v}}_i,\widetilde{\bm{q}}_j\}}\![g(\mathcal{S}\! \cup\! \{\widetilde{\bm{v}}_i,\widetilde{\bm{q}}_j\})\! 
+\! g(\mathcal {S})\!-\! g(\mathcal{S}\! \cup\! \{\widetilde{\bm{v}}_i\})\!-\! g(\mathcal{S}\! \cup\! \{\widetilde{\bm{q}}_j\})],
\label{banzhafinter}
\end{align}\normalsize
where $\frac{1}{2^{n-2}}$ is the likelihood of $\mathcal {S} \subseteq  \mathcal{P} \backslash  \{\widetilde{\bm{v}}_i,\widetilde{\bm{q}}_j\}$ being sampled. 
Intuitively, $I_B(\{\widetilde{\bm{v}}_i,\widetilde{\bm{q}}_j\})$ quantifies this additional benefit compared to when players $\widetilde{\bm{v}}_i$ and $\widetilde{\bm{q}}_j$ work independently. It also embodies the tendency of interaction between  frame-player and word-player. A higher value of $I_B(\{\widetilde{\bm{v}}_i,\widetilde{\bm{q}}_j\})$ signifies that players $\widetilde{\bm{v}}_i$ and $\widetilde{\bm{q}}_j$ cooperate closely with each other, and the formed coalition will bring additional high returns. By this interaction index, we can measure the closeness between plays $\widetilde{\bm{v}}_i$ and $\widetilde{\bm{q}}_j$ to achieve fine-grained alignment between video and query. 

\noindent \textbf{Game details.}
To ensure complete consistency between the cooperative game and cross-modal alignment learning, the game function $g$ should meet the following criteria: 1) the game payoff benefits from strongly corresponding semantic pairs $\{\bm{v}_i^+,\bm{q}_j^+\}$, \textit{i.e.}, $g(\mathcal{P}) - g(\mathcal{P}\backslash \{\bm{v}_i^+,\bm{q}_j^+\} \cup \{[\{\bm{v}_i^+,\bm{q}_j^+\}] \})<0$; 2) the game payoff is compromised by semantically opposite pairs $\{\bm{v}_i^-,\bm{q}_j^-\}$, \textit{i.e.}, $g(\mathcal{P}) - g(\mathcal{P}\backslash \{\bm{v}_i^-,\bm{q}_j^-\} \cup \{[\{\bm{v}_i^-,\bm{q}_j^-\}] \})>0$;
3) when there are no players to cooperate, the payoff is zero, \textit{i.e.},  $g(\{ \bm{v}_i\}_{i=1}^{T}) = g(\{ \bm{q}_j\}_{j=1}^{M}) =  g(\phi) = 0$, where $\phi$ is the empty set.
It should be noted that any function that satisfies the aforementioned conditions can be employed as the game function $g(\cdot)$. To simplify matters, we utilize the cosine similarity as $g$ in all games. 

During the game, when players $\bm{v}_i$ and $\bm{q}_j$ form a coalition $\{\bm{v}_i,\bm{q}_j\}$, we deem $[\{\bm{v}_i,\bm{q}_j\}]$ as a singular hypothetical player, which is the union of the player in $\{\bm{v}_i,\bm{q}_j\}$. Then, the reduced game is formed by removing the individual players in $\{\bm{v}_i,\bm{q}_j\}$ from the game and adding $[\{\bm{v}_i,\bm{q}_j\}]$ to the game.

\noindent \textbf{Word-level alignment with local game interaction.} 
Similar to the self-modal game,
we develop an alignment header to predict the word-frame alignment matrix $A = [a_{i,j}]^{T \times M}$ between frame $\widetilde{\bm{v}}_i$ and word $\widetilde{\bm{q}}_j\}$, supervised by $I_B(\{\widetilde{\bm{v}}_i,\widetilde{\bm{q}}_j\})$. 
Then, we optimize the Kullback-Leibler Divergence (KLD) between the $I_B(\{\widetilde{\bm{v}}_i,\widetilde{\bm{q}}_j\})$ and $a_{i,j}$, to bring the probability distributions of the output of the alignment header and interaction  index close together to establish word-level semantic alignment between frame players and word players.

In detail, 
we define the word-to-frame probability distribution $d^j_{w2f}$ of the $j$-th word-to-frame alignment as $d^j_{w2f}\!=\![p_{1,j},...,p_{t,j}]$, where $p_{i,j}\!=\! \frac{\exp{(I_B(\{\widetilde{\bm{v}}_i,\widetilde{\bm{q}}_j\}))}}{\sum_{l=1}^T \exp{(I_B(\{\widetilde{\bm{v}}_l,\widetilde{\bm{q}}_j\}))}}$, and the total query-to-video probability distribution $D_{q2v} = [d^1_{w2f},...,d^M_{w2f}]^\top$.
Similarly, 
the predicted fine-grained probability distribution can be represented as $\hat{d}^j_{w2f} \!=\![\hat{p}_{1,j},...,\hat{p}_{t,j}]$, where $\hat{p}_{i,j} = \frac{\exp{(a_{i,j})}}{\sum_{l=1}^t \exp{ (a_{l,j})}}$, and the total predicted probability distribution is defined as $\hat{D}_{q2v} = [\hat{d}^1_{w2f},...,\hat{d}^M_{w2f}]^\top$. At last, the word-level alignment loss $\mathcal{L}_{word}$ is formulated as:
\small
\begin{align}
\mathcal{L}_{word} ={\rm KLD} (\hat{D}_{q2v}||{D}_{q2v})
= -\frac{1}{TM}\sum\nolimits_{i=1}^{T}\sum\nolimits_{j=1} ^{M} {p}_{i,j}  (\log{\hat{p}_{i,j}} -\log{{p}_{i,j}} ).
\label{giloss}
\end{align}\normalsize





\noindent \textbf{Phrase- and sentence-level alignment with global game interaction.}
Although the above cross-modal game-theoretic interaction between each possible player-pair $\{\widetilde{\bm{v}}_i,\widetilde{\bm{q}}_j\}$ addresses the problem of word-level alignment between  word player $\widetilde{\bm{v}}_i$ and frame player $\widetilde{\bm{q}}_j$, 
it is still limited to aligning frames with isolated words, and fails to explore the relations between frames with more contextual phrase- or sentence-level semantics for better understanding temporal events.
To this end, we introduce a multi-level game interaction to measure the semantic consistency between frames with different-grained textual semantics.
Specifically, we follow  \cite{liu2020jointly,mun2020local} to utilize Sequential Query Attention Network (SQAN) to extract the phrase-level textual semantics $\widetilde{\bm{Q}}^{p}=\{\widetilde{\bm{q}}^p_j\}_{j=1}^P$, and $P$ is the phrase number. The sentence-level semantics $\widetilde{\bm{Q}}^{s}$ is obtained by concatenating the last hidden states in both the forward and backward BiLSTM. 
Similar to worl-level alignment, we compute both phrase- and sentence-level text-to-frame game interaction as $I_B(\{\widetilde{\bm{v}}_i,\widetilde{\bm{q}}_j^p\}),I_B(\{\widetilde{\bm{v}}_i,\widetilde{\bm{Q}}^s\})$ via \cref{banzhafinter}. We design corresponding alignment headers (phrase-level cross-modal header and sentence-level cross-modal header) to predict their alignment matrix via $\mathcal{L}_{phra},\mathcal{L}_{sent}$.

\subsection{Model Training and Inference}
\noindent \textbf{Weak supervision for video-query matching.}
We first interact each frame with the semantics of words, phrases and the whole sentence to calculate the overall query-to-frame matching degree by:
\small
\begin{equation}
m_i = \frac{1}{3} (I_B(\{\widetilde{\bm{v}}_i,\widetilde{\bm{Q}}^s\}) + \frac{1}{M}\sum\nolimits_{j=1}^{M}I_B(\{\widetilde{\bm{v}}_i,\widetilde{\bm{q}}_j\})+\frac{1}{P}\sum\nolimits_{l=1}^{P}I_B(\{\widetilde{\bm{v}}_i,\widetilde{\bm{q}}_l^p\})),
\label{matching}
\end{equation}\normalsize
where 
$m_i $ is the overall query-to-frame matching degree of $i$-th frame. 

Then, the whole video-sentence matching score is calculated by:
\small
\begin{equation}
\phi^c=\sum\nolimits_{i=1}^T w_i\cdot m_i,  \ w_i=\frac{\exp{(I_B(\{\widetilde{\bm{v}}_i,\widetilde{\bm{Q}}^s\}))}}{\sum_{i=1}^{T}\exp (I_B(\{\widetilde{\bm{v}}_i,\widetilde{\bm{Q}}^s\}))},
\label{simlarity}
\end{equation}\normalsize
where we utilize the sentence-level similarity to represent the contribution of each frame to the target moment among the entire video.
For matched video $V_k$ and query $Q_k$,
we calculate the video-sentence matching score $\phi^c_{V_k,Q_k}$. The scores of unmatched pairs are expressed as $\phi^c_{V_k,Q_l}$ and $\phi^c_{Q_k,V_l}$ where $l\neq k$. 

Finally, we utilize the cross-modal contrastive loss~\cite{oord2018representation} for weak supervision:
\small
\begin{align}
\mathcal{L}_{C}\!=\! - \frac{1}{N_b}\!\sum\nolimits_{k=1}^{N_b}\!\log{\frac{\exp{(\phi^c_{V_k,Q_k}/\tau)}}{\sum_{l=1}^{N_b}\exp({\phi^c_{V_k,Q_l}}/\tau)}}\!
-\!\frac{1}{N_b}\sum\nolimits_{k=1}^{N_b}\log{\frac{\exp{(\phi^c_{Q_k,V_k}/\tau)}}{\sum_{l=1}^{N_b}\exp({\phi^c_{Q_k,V_l}}/\tau)}},
\label{conloss}
\end{align}\normalsize
where $\tau$ is the temperature parameter, ${N_b}$ is the batch size. 

\noindent \textbf{Reconstruction loss.} 
To further enhance the semantic relevance of the highlighted frames to the sentence, we reconstruct the query conditioned on highlighted frames. 
Following popular reconstruction strategy \cite{lin2020weakly,zheng2022cnm}, we randomly replace the 1/3 words in the original query with a specific symbol, and predict the next word by given a prefix of the query and highlighted frames feature. Specifically, we embed the masked query $\bm{Q}^m$ by Glove and predict the next word $\hat{\bm{q}}_{i+1}$ using the conditioned transformer following CNM~\cite{zheng2022cnm} with input $(\bm{\widetilde{V}},C_m,\bm{Q}^m_{1:i})$, where $C_m = \{m_1,m_2,...,m_T\}$ represents the highlighted frame mask, $\bm{Q}^m_{1:i}$ denotes the masked query words from the $1$-st to $i$-th. 

Then, we introduce the cross-entropy loss $\mathcal{L}_{rec}$ to calculate the probability distribution $\mathcal{P}^q$ different between the predict word $\hat{\bm{q}}_{i+1}$ and the real word as:
\small
\begin{equation}
\mathcal{L}_{rec}= - \sum\nolimits_{i=1}^{M-1}\log{\mathcal{P}^q(\bm{q}_{i+1}|\hat{\bm{q}}_{i+1}}).
\label{recloss}
\end{equation}\normalsize


\noindent \textbf{Overall training.}
To train the whole model, we combine game-guided self-relevance losses $\mathcal{L}_{self}=\mathcal{L}_{self}^v+ \mathcal{L}_{self}^q$, game alignment loss $\mathcal{L}_{align}=\mathcal{L}_{word}+ \mathcal{L}_{phra}+ \mathcal{L}_{sent}$, video-level contrastive loss $\mathcal{L}_{C}$, and reconstruction loss $\mathcal{L}_{rec}$ with trade-off hyper-parameters ($\alpha_1, \alpha_2, \alpha_3$ and $\alpha_4$) to define the overall loss as:
\small
\begin{equation}
\mathcal{L}= \alpha_1 \mathcal{L}_{self}+\alpha_2\mathcal{L}_{align}+\alpha_3 \mathcal{L}_{C}+ \alpha_4 \mathcal{L}_{rec}.
\label{allloss}
\end{equation}\normalsize

\noindent \textbf{Inference.}
During inference, we directly utilize the self-relevance headers to enhance the video and query features, and exploit the multi-level alignment headers to predict the overall query-to-frame matching degree. We locate the frame with the highest score as the basic predicted moment, and add the left/right frames into the moment if the ratio of their scores to the frame score of the closest moment boundary is higher than a threshold. We repeat this step until no frame can be added. 
We can obtain more moments  by locating different initial frames with different scores. 


\section{Experiments}
\label{others}

\subsection{Datasets and Evaluation Metrics}
\noindent \textbf{Charades-STA.}
The Charades-STA dataset \cite{gao2017tall} is built based on the Charades \cite{sigurdsson2016hollywood}, which contains 6,672 videos of indoor activities and involves 16,128 query-video pairs. There are 12,408 pairs used for training and 3,720 used for testing. The average duration of each video is 29.76 seconds.

\noindent \textbf{ActivityNet Caption.}
The ActivityNet Caption dataset \cite{krishna2017dense} contains 20,000 videos with 100,000 queries, where 37,421 query-video pairs are used for training and 34,536 are used for testing. 
On average, each video in ActivityNet Caption has 3.65 annotated moments and each annotated moment lasts for 36 seconds.
For  fair comparison, we follow \cite{mithun2019weakly,chen2020look,song2020weakly,tan2021logan,zhang2020counterfactual,zheng2022cnm} to use val1 as valuation and report the test performance on val2.

\noindent \textbf{Evaluation metrics.}
Following previous works,
we adopt the metrics ``R@$n$, IoU=$m$'' to evaluate our model, which presents the proportion of the top $n$ predicted moments with IoU larger than $m$. 

\subsection{Implementation Details}
For fair comparison,
we follow previous works to apply C3D~\cite{tran2015learning} to encode the videos on ActivityNet Caption and I3D~\cite{carreira2017quo} on Charades-STA. 
We uniformly downsample the length of video feature sequences to $T=200$ for ActivityNet Caption and $T=64$ for Charades-STA. 
As for query encoding, we also follow previous works to utilize set the length of word feature sequences to $M=20$, and utilize Glove~\cite{pennington2014glove} to embed each word.
The dimension $D_h$ is set to 512. These trade-off hyper-parameters are set to
$\alpha_1=1.0,\alpha_2=1.0,\alpha_3=1.0,\alpha_4=10$.
Since the calculation of the exact game-theoretic interaction is an NP-hard problem~\cite{matsui2001np},
we follow previous works to utilize sampling-based method \cite{bachrach2010approximating,leech2002computation} to obtain unbiased estimates for approximating it. 
We train the whole model for 100 epochs with batch size of 16 and early stopping strategy. Parameter optimization is performed by Adam~\cite{kingma2014adam} optimizer with learning rate $3\times 10^{-4}$, and linear decay of learning rate and gradient clipping of 1.0.
The inference threshold is set to 0.8 in ActivityNet Caption and 0.9 in Charades-STA.

\subsection{Main Results}
\noindent \textbf{Quantitative comparison.}
We compare our method with the state-of-the-art methods, including 
\textit{fully-supervised} (FS) methods and \textit{weakly-supervised} (WS) methods. 
Best results are in \textbf{bold}. 
We introduce two types of cooperative game indices, \textit{i.e.}, Banzhaf index \cite{grabisch1999axiomatic}  and Shapley index \cite{queshapley1953value}, to our model. As summarized in \cref{tab:comparison}, both two different variants of our method surpass all existing methods on both Charades-STA and ActivityNet Captions. Particularly, on Charades-STA in terms of ``R@5, IoU=0.5'', Ours(Banzhaf) outperforms the best compared method CPL by 4.53\%. The main reason is that our game-theoretic-based approach can learn the local alignment between words and frames for accurately predicting the moment boundaries.
Besides, the Shapley variant performs similar to the Banzhaf variant for this specific WS-VTG task, since they share the same game interaction formulation with different normalization.
Overall, this table verifies the effectiveness of our game-based framework.

\begin{table*}[t!]
\small
\centering
\caption{\small{\textbf{Comparisons with  state-of-the-art methods.} ``WS'' and ``FS'' denotes  weakly-supervised and fully-supervised methods, respectively. 
}}
\vspace{-12pt}
\scalebox{0.88}{
\setlength{\tabcolsep}{0.5mm}{
\begin{tabular}{cccccccccc}
\toprule
\multirow{3}{*}{Type}&\multirow{3}{*}{Method}& \multicolumn{4}{c}{Charades-STA} & \multicolumn{4}{c}{ActivityNet Caption}  \\\cmidrule(lr){3-6} \cmidrule(lr){7-10}
& & R@1,  & R@1,  & R@5,  & R@5,  & R@1, & R@1,       & R@5,       & R@5,  \\
& & IoU=0.5 &IoU=0.7 & IoU=0.5 &IoU=0.7 & IoU=0.3 &IoU=0.5 & IoU=0.3 &IoU=0.5
 \\ \midrule

\multirow{3}{*}{FS}&CTRL~\cite{gao2017tall}         & 23.63       & 8.89  & 58.92   & 29.52     & - & 29.01       & -       & 59.17     \\

&2DTAN~\cite{Alpher33}            & 39.81 &23.25 &79.33 &51.15 &  59.45      & 44.51 & 85.53  & 77.13    \\
&DRN~\cite{zeng2020dense}             & 53.09 &31.75& 89.06& 60.05      &- &45.45 &- &77.97

 \\ \midrule
\multirow{13}{*}{WS}& SCN~\cite{lin2020weakly} &23.58& 9.97& 71.80& 38.87& 47.23 &29.22 &71.45& 55.69\\
&WSTAN~\cite{wang2021weakly}& 29.35 &12.28&76.13&41.53&52.45& 30.01 &79.38& 63.42\\
&ICVC~\cite{chen2022explore}&31.02& 16.53& 77.53& 41.91& 46.62 &29.52& 80.92 &66.61\\
&MARN \cite{song2023marn}& 33.87& 15.54&  73.90& 41.94& 48.52& 31.37&75.91& 60.00\\
&CRM~\cite{huang2021cross}&34.76& 16.37& -& -& 55.26& 32.19&-&- \\
&CNM~\cite{zheng2022cnm}& 35.15 & 14.95 & -& -&55.68 &33.33 &-&-\\
&ACN \cite{wu2023atomic}& 37.02& 15.26& -&-& 57.66& 34.18 & -&-\\
&VCA~\cite{wang2021visual}&38.13&19.57& 78.75& 37.75& 50.45& 31.00& 71.79 &53.83\\
&LCNet~\cite{yang2021local} &39.19& 18.17 &80.56 &45.24 &48.49& 26.33& 82.51 &62.66
\\
&CPL~\cite{zheng2022cpl} &49.24 &22.39& 84.71& 52.37 &55.73& 31.37 &63.05 &43.13\\
&DM2 \cite{ma2023dual}& 51.39& 23.72& -&-&  54.73& 31.85&-&-\\
\cmidrule{2-10}
&\textbf{Ours(Banzhaf)}  & 54.17 & \textbf{26.46} & 88.93 & \textbf{55.79} & 58.94 & 35.65 & 86.41 & \textbf{70.26} \\
&\textbf{Ours(Shapley)}  & \textbf{54.31} & 26.28 & \textbf{89.24} & 55.62 &\textbf{59.17} & \textbf{35.80} & \textbf{86.59} & 70.03  \\
\bottomrule
\label{tab:comparison}
\end{tabular}}}
\vspace{-5mm}
\end{table*}
\begin{figure}[t!]
\centering
\begin{minipage}[t]{0.48\textwidth}
        \makeatletter\def\@captype{table}\makeatother
\footnotesize
\centering
\setlength\tabcolsep{2pt}
\caption{\small{\textbf{Cross-dataset evaluation.} ``$A\rightarrow C$'' denotes training on ActivityNet Caption and evaluation on Charades-STA, and vice versa.}}
\vspace{-10pt}
\scalebox{0.76}{
\setlength{\tabcolsep}{0.5mm}{
\begin{tabular}{ccccc}
\toprule
\multirow{3}{*}{Method}& \multicolumn{2}{c}{ $A\rightarrow C$} & \multicolumn{2}{c}{$C\rightarrow A$} \\\cmidrule(lr){2-3} \cmidrule(lr){4-5}
&   R@1,    & R@5,  & R@1,         & R@5,\\ 
&   IoU=0.7    & IoU=0.7   & IoU=0.5         & IoU=0.5\\
\midrule
LCNet~\cite{yang2021local} & 12.06 & 31.47 & 10.65 & 31.76 \\ 
CPL~\cite{zheng2022cpl} & 14.32 & 35.81 & 11.59 & 29.88 \\
\textbf{Ours(Banzhaf)} & \textbf{18.74} & 42.58 & \textbf{20.37} & \textbf{42.96} \\
\textbf{Ours(Shapley)} & 18.26 & \textbf{43.13} & 20.04 & 41.42
\\
\bottomrule
\label{tab:ablation3}
\end{tabular}}}
\end{minipage}
   \hspace{0.1in}
   \begin{minipage}[t]{0.48\textwidth}
\centering
 \vspace{-2mm}
	\includegraphics[width=\linewidth]{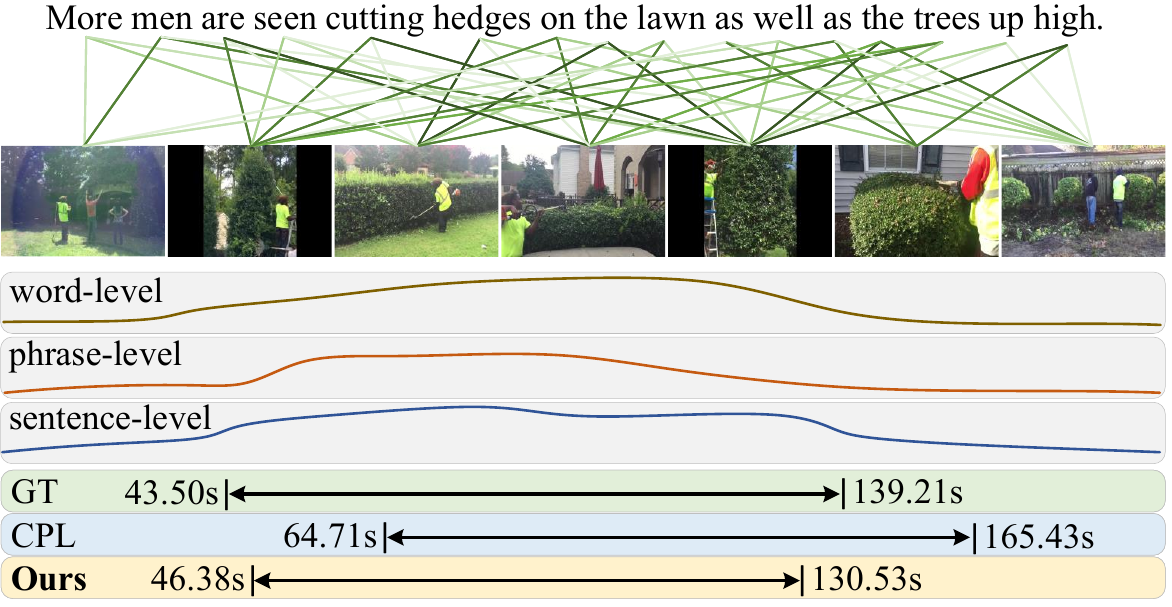}
	\vspace{-6mm}
\caption{\small{
	\textbf{Qualitative results.} The colors of green lines represent different degrees of confidence.}
	}
	\label{fig:result}
 \end{minipage}
 \vspace{-9mm}
\end{figure}

\noindent \textbf{Generalization on cross-dataset evaluation.}
At last, we explore the generalization of models by cross-dataset evaluation in \cref{tab:ablation3}. Concretely, we first train a model on one dataset and then evaluate its performance on the other dataset. We can find that: 1) For all WS-VTG methods, the performance of A $\rightarrow$ C achieves better than C $\rightarrow$ A. This is because that ActivityNet Caption is larger and more complex than Charades-STA, thus the former is able to bring more generalized knowledge to the latter while the latter fails to provide general knowledge to the former.
2) Our two variants outperform  state-of-the-art methods (\eg, LCNet) with clear margins on cross-dataset evaluation. We speculate it to: Previous works severely rely on the moment proposals, however, LCNet defines different sliding windows on different datasets for proposal generation. Therefore, its non-uniform proposal structure limits its generalization on cross-dataset evaluation. Instead, we learn multi-level frame-aware alignment and utilize frame-wise scores for predicting moment, which is more flexible and adaptable, thus improving the generalization-ability.

\noindent \textbf{Qualitative comparison.}
In \cref{fig:result}, we provide a qualitative example  on ActivityNet
Caption. 
The colors of green lines represent different degrees of interaction confidence, demonstrating that the game interaction can well learn the uncertain and fine-grained frame-word correspondence in the WS setting.
Besides, all three-level interaction (word-level, phrase-level and sentence-level) contributes differently to the grounding, and our final method is able to localize more accurate moment boundaries than state-of-the-art method CPL. 

\begin{table}[t!]
    \small
    \centering
    \caption{\small  Comparison on speed per sample (s) and GPU memory cost (MB).}
    \vspace{-10pt}
    \setlength{\tabcolsep}{1.2mm}{
    \begin{tabular}{c|c|cccc}
    \hline
    Metric & Stage & LCNet \cite{yang2021local}  & CPL \cite{zheng2022cpl} & Ours(Banzhaf) & Ours(Shapley) \\ \hline
    \multirow{2}*{Speed} & Training & 0.79s & \textbf{0.27s} & 0.94s & 0.98s \\
    ~ & Inference & 0.62s & 0.21s & \textbf{0.10s} & 0.11s\\ \hline
    GPU & Training  & 8647MB & \textbf{4156MB} & 5237MB & 5482MB \\
    Memory&  Inference & 7462MB & 3428MB & \textbf{2058MB} & 2165MB \\ \hline
    \end{tabular}}
    \vspace{-5pt}
    \label{tab:efficient}
\end{table}

\noindent \textbf{Complexity comparison.}
In \cref{tab:efficient}, our method costs more training time/memory as we need to enumerate coalition samples. Specifically, we sampled 5500 samples. We find that, the model instability decreases along with the increase of the sampling number, and when the sample size ranges from 4500 to 5500, the error with the true value of banzhaf gradually decreases from 6.9\% to 1.1\%. After exceeding 5500 samples, more samples will not affect the result. 
During inference, since we can directly utilize the learned game headers to generate game value instead of using complex game learning, our model is faster with lower memory.

\begin{table*}[!t]
\small
\centering
\setlength\tabcolsep{0.8pt}
\caption{\small\textbf{Main ablation study.} }
\vspace{-8pt}
\scalebox{0.74}{
\begin{tabular}{ccccccccccccccc}
\toprule
\multicolumn{2}{c}{Self-modal} & \multicolumn{3}{c}{Multi-level} & \multicolumn{4}{c}{Banzhaf-based Game}&\multicolumn{4}{c}{Shapley-based Game}\\\cmidrule(lr){6-9} \cmidrule(lr){10-13}
\multicolumn{2}{c}{Game} & \multicolumn{3}{c}{Cross-modal Game}&\multicolumn{2}{c}{Charades-STA} & \multicolumn{2}{c}{ActivityNet Caption} &\multicolumn{2}{c}{Charades-STA} & \multicolumn{2}{c}{ActivityNet Caption}\\\cmidrule(lr){1-2} \cmidrule(lr){3-5} \cmidrule(lr){6-7} \cmidrule(lr){8-9} \cmidrule(lr){10-11} \cmidrule(lr){12-13}
Video & Query & Word- & Phrase- & Sentence-   & R@1,    & R@5,   & R@1,         & R@5, & R@1,    & R@5,   & R@1,         & R@5, \\
Game& Game & level & level & level &IoU=0.7  &IoU=0.7  &IoU=0.5  &IoU=0.5 &IoU=0.7  &IoU=0.7  &IoU=0.5  &IoU=0.5 \\ \midrule
$\times$ & $\times$ & $\checkmark$ & $\times$ & $\times$ & 19.34 & 44.67 & 29.83 & 61.72 & 19.29 & 44.31 & 30.15 & 61.56 \\
$\checkmark$ & $\times$ & $\checkmark$ & $\times$ & $\times$ & 21.25 & 47.86 & 31.39 & 63.98 & 21.10 & 47.73 & 31.54 & 63.67 \\
$\checkmark$ & $\checkmark$ & $\checkmark$ & $\times$ & $\times$ & 22.97 & 50.64 & 32.72 & 65.80 & 22.69 & 50.38 & 32.97 & 65.65 \\
$\checkmark$ & $\checkmark$ & $\checkmark$ & $\checkmark$ & $\times$ & 24.81 & 53.28 & 34.46 & 68.03 & 24.64 & 53.03 & 34.59 & 67.82 \\
$\checkmark$ & $\checkmark$ & $\checkmark$ & $\checkmark$ & $\checkmark$ & \textbf{26.46} & \textbf{55.79} & \textbf{35.65} & \textbf{70.26} & \textbf{26.28} & \textbf{55.62} & \textbf{35.80} & \textbf{70.03} \\
\bottomrule
\label{tab:ablation}
\end{tabular}}
\vspace{-8mm}
\end{table*}
\subsection{Ablation Study}
In this section, we perform in-depth ablation studies to evaluate the effect of each component. 
We implement our model with different game settings.

\noindent \textbf{Main ablation.}
As shown in \cref{tab:ablation}, we conduct ablation studies regarding the components (\textit{i.e.}, self-modal games in two modalities, and  multi-level cross-modal games). We observe the following  findings:
1) Our baseline model (only contains word-level cross-modal game) is powerful and outperforms some  WS methods, demonstrating that it is effective to capture the uncertain yet fine-grained alignment for better determining the boundaries.
2) Both video and query self-modal games bring significant improvement to the overall model, indicating that self-enhancement is crucial for context integration.
3) Each query level of cross-modal game further boost the performance, verifying that phrase- and sentence-level semantics help to better understand temporal events.
Overall, all components contribute a lot to  final performance, showing their effectiveness.

\begin{table*}[!t]
\small
\centering
\setlength\tabcolsep{1pt}
\caption{\small{\textbf{Ablation study on each component}}, where ``LL'' means ``Linear Layer'', ``CL'' means ``Convolution Layer'', ``SA'' means ``Self-attention''; ``MSE'' means ``Mean Squared Error'', ``CE'' means ``Cross Entropy'', ``KLD'' means ``Kullback-Leibler Divergence''; ``RL'' means ``Reconstruction Loss''.}
     \vspace{-12pt}
\scalebox{0.86}{
\begin{tabular}{lcccccccccccc}
\toprule
\multirow{4}{*}{Component}&\multirow{4}{*}{Setting}& \multicolumn{4}{c}{Banzhaf-based Game}& \multicolumn{4}{c}{Shapley-based Game}\\\cmidrule(lr){3-6} \cmidrule(lr){7-10}
&&\multicolumn{2}{c}{Charades-STA} & \multicolumn{2}{c}{ActivityNet Caption}&\multicolumn{2}{c}{Charades-STA} & \multicolumn{2}{c}{ActivityNet Caption}\\\cmidrule(lr){3-4} \cmidrule(lr){5-6}\cmidrule(lr){7-8} \cmidrule(lr){9-10}
&   & R@1,    & R@5,   & R@1,         & R@5, & R@1,    & R@5,   & R@1,         & R@5,  \\
&  &IoU=0.7  &IoU=0.7  &IoU=0.5  &IoU=0.5 &IoU=0.7  &IoU=0.7  &IoU=0.5  &IoU=0.5
 \\ \midrule
\multirow{4}{*}{\makecell{Game \\ Headers}}& LL & 25.28 & 53.90 & 34.15 & 68.03 & 24.97 & 53.68 & 34.25 & 67.94\\
& CL & 25.52 & 54.31 & 34.56 & 68.64 & 25.26 & 54.03 & 34.82 & 68.47 \\
 & LL + SA & 25.94 & 54.83 & 35.27 & 69.18 & 25.71 & 54.59 & 35.53 & 69.06\\
& CL + SA &  \textbf{26.46} & \textbf{55.79} & \textbf{35.65} & \textbf{70.26} &  \textbf{26.28} & \textbf{55.62} & \textbf{35.80} & \textbf{70.03} \\ \midrule
 \multirow{4}{*}{\makecell{Soft \\ Supervision}} & None & 23.21 & 51.64 & 32.36 & 66.40 & 22.99 & 51.45 & 32.71 & 66.18\\
 & MSE & 24.42 & 53.19 & 33.85 & 67.57 & 24.24 & 52.89 & 40.06 & 67.33\\
 & CE & 24.73 & 53.58 & 34.16 & 68.04 & 24.52 & 53.26 & 34.37 & 67.80\\
 & KLD & \textbf{26.46} & \textbf{55.79} & \textbf{35.65} & \textbf{70.26} & \textbf{26.28} & \textbf{55.62} & \textbf{35.80} & \textbf{70.03}\\\midrule
\multirow{2}{*}{\makecell{Grounding \\ Losses}}& w/o RL & 24.35 & 53.57 & 34.14 & 68.42 & 24.08 & 53.31 & 34.45 & 68.19\\
& w/ RL & \textbf{26.46} & \textbf{55.79} & \textbf{35.65} & \textbf{70.26} & \textbf{26.28} & \textbf{55.62} & \textbf{35.80} & \textbf{70.03}\\
\bottomrule
\label{tab:ablation2}
\end{tabular}}
\vspace{-8mm}
\end{table*}

\noindent \textbf{Effect of  game headers.} The headers in both self- and cross-modal games are utilized to provide soft supervision for better game interaction and player-representation learning. To explore the impact of the structure of these headers, we compare several popular structures in \cref{tab:ablation2}. We find that the combination of convolution layer and self-attention can capture both local and global interaction, so it achieves the best performance and is beneficial for cross-modal multi-level interaction, illustrating the significance of our game headers.

\noindent \textbf{Necessity of  soft supervision.} To explore the effect of soft supervision, we conduct the ablation study to investigate the contribution of the KLD loss during the header learning in \cref{tab:ablation2}. The KLD loss achieves better performance than both MSE and CE, showing that it is more suitable for game headers to mimic game-theoretic interaction learning. Moreover, without soft supervision, the performance will drop a lot, showing the importance of the soft supervision.

\noindent \textbf{Analysis on reconstruction loss.}
Most previous WS-VTG works also additionally utilize the reconstruction paradigm to assist the model training. As shown in \cref{tab:ablation2}, we further conduct such ablation study to investigate the impact of reconstruction loss. It shows that the reconstruction loss can improve significantly the performance, demonstrating its effectiveness.

\section{Conclusion}
In this paper, we rethink the limitations of previous weakly-supervised video temporal grounding works, and creatively propose to utilize the game-theoretic interaction to learn the uncertain relationship between video-query pairs with diverse granularity.
Specifically, we first introduce two self-modal games to correlate the frames/words to enhance their contextual semantics. Then, we propose a cross-modal game to value fine-grained correspondence between frames and words with multiple levels for better determining the query-related moment boundaries.
Extensive experiments demonstrate the effectiveness of our game-based framework on two challenging datasets.

\bibliographystyle{splncs04}
    \bibliography{reference,full}

@article{dong2022dual,
  title={Dual encoding for video retrieval by text},
  author={Dong, Jianfeng and Li, Xirong and Xu, Chaoxi and Yang, Xun and Yang, Gang and Wang, Xun and Wang, Meng},
  journal={IEEE Transactions on Pattern Analysis and Machine Intelligence},
  volume={44},
  number={8},
  pages={4065--4080},
  year={2022},
  publisher={IEEE}
}

@article{fang2025your,
  title={Your data is not perfect: Towards cross-domain out-of-distribution detection in class-imbalanced data},
  author={Fang, Xiang and Easwaran, Arvind and Genest, Blaise and Suganthan, Ponnuthurai Nagaratnam},
  journal={Expert Systems with Applications},
  year={2025}
}

@article{fang2023hierarchical,
  title={Hierarchical local-global transformer for temporal sentence grounding},
  author={Fang, Xiang and Liu, Daizong and Zhou, Pan and Xu, Zichuan and Li, Ruixuan},
  journal={IEEE Transactions on Multimedia},
  year={2023},
  publisher={IEEE}
}

@article{fang2022multi,
  title={Multi-modal cross-domain alignment network for video moment retrieval},
  author={Fang, Xiang and Liu, Daizong and Zhou, Pan and Hu, Yuchong},
  journal={IEEE Transactions on Multimedia},
  volume={25},
  pages={7517--7532},
  year={2022},
  publisher={IEEE}
}

@inproceedings{fang2026cogniVerse,
  title={CogniVerse: Revolutionizing Multi-modal Retrieval-Augmented Generation with Cognitive Reflection and Geometric Reasoning},
  author={Fang, Xiang and Fang, Wanlong and Wang, Changshuo},
  booktitle={Proceedings of the IEEE/CVF Conference on Computer Vision and Pattern Recognition},
  year={2026}
}

@inproceedings{fang2023you,
  title={You can ground earlier than see: An effective and efficient pipeline for temporal sentence grounding in compressed videos},
  author={Fang, Xiang and Liu, Daizong and Zhou, Pan and Nan, Guoshun},
  booktitle={Proceedings of the IEEE/CVF Conference on Computer Vision and Pattern Recognition},
  pages={2448--2460},
  year={2023}
}

@inproceedings{fang2025hierarchical,
  title={Hierarchical Semantic-Augmented Navigation: Optimal Transport and Graph-Driven Reasoning for Vision-Language Navigation},
  author={Fang, Xiang and Fang, Wanlong and Wang, Changshuo},
  booktitle={Advances in Neural Information Processing Systems},
  year={2025}
}

@inproceedings{fang2025adaptive,
  title={Adaptive Multi-prompt Contrastive Network for Few-shot Out-of-distribution Detection},
  author={Fang, Xiang and Easwaran, Arvind and Genest, Blaise},
  booktitle={International Conference on Machine Learning},
  year={2025}
}

@inproceedings{fang2026slap,
  title={SLAP: The Semantic Least Action Principle for Variational Video-Language Modeling},
  author={Fang, Xiang and Fang, Wanlong},
  booktitle={International Conference on Machine Learning},
  year={2026}
}

@inproceedings{fang2026immuno,
  title={Immuno-VLM: Immunizing Large Vision-Language Models via Generative Semantic Antibodies for Open-World Trustworthiness},
  author={Fang, Xiang and Fang, Wanlong and Ji, Wei},
  booktitle={International Conference on Machine Learning},
  year={2026}
}

@inproceedings{fang2026disentangling,
  title={Disentangling Adversarial Prompts: A Semantic-Graph Defense for Robust LLM Security},
  author={Fang, Xiang and Fang, Wanlong},
 booktitle={Proceedings of the AAAI Conference on Artificial Intelligence},
year={2026}
}

@inproceedings{fang2026advancing,
  title={Advancing Out-of-Distribution Detection Across Diverse Scenarios},
  author={Fang, Xiang},
  booktitle={Proceedings of the AAAI Conference on Artificial Intelligence},
  volume={40},
  number={48},
  pages={41042--41043},
  year={2026}
}

@inproceedings{fang2026unveiling,
  title={Unveiling the Fragility of Vision-Language Models: Multi-Modal Adversarial Synergy via Texture-Constrained Perturbations and Cross-Modal Optimization},
  author={Fang, Xiang and Fang, Wanlong and Wang, Changshuo},
 booktitle={Proceedings of the AAAI Conference on Artificial Intelligence},
year={2026}
}

@inproceedings{fang2026rethinking,
  title={Rethinking Video-language Model From the Language Input Perspective},
  author={Fang, Xiang and Fang, Wanlong and Wang, Changshuo and Qu, Xiaoye and Liu, Daizong},
 booktitle={Proceedings of the AAAI Conference on Artificial Intelligence},
year={2026}
}

@inproceedings{fang2026towards,
  title={Towards Unified Vision-Language Models With Incomplete Multi-Modal Inputs},
  author={Fang, Xiang and Fang, Wanlong and Wang, Changshuo and Tang, Keke and Liu, Daizong and Wang, Siyi and Ji, Wei},
 booktitle={Proceedings of the AAAI Conference on Artificial Intelligence},
year={2026}
}

@inproceedings{fang2025multi,
  title={Multi-pair temporal sentence grounding via multi-thread knowledge transfer network},
  author={Fang, Xiang and Fang, Wanlong and Wang, Changshuo and Liu, Daizong and Tang, Keke and Dong, Jianfeng and Zhou, Pan and Li, Beibei},
  booktitle={Proceedings of the AAAI Conference on Artificial Intelligence},
  volume={39},
  number={3},
  pages={2915--2923},
  year={2025}
}

@inproceedings{fang2024fewer,
  title={Fewer Steps, Better Performance: Efficient Cross-Modal Clip Trimming for Video Moment Retrieval Using Language},
  author={Fang, Xiang and Liu, Daizong and Fang, Wanlong and Zhou, Pan and Xu, Zichuan and Xu, Wenzheng and Chen, Junyang and Li, Renfu},
  booktitle={Proceedings of the AAAI Conference on Artificial Intelligence},
  volume={38},
  number={2},
  pages={1735--1743},
  year={2024}
}

@inproceedings{fang2024multi,
  title={Multi-Pair Temporal Sentence Grounding via Multi-Thread Knowledge Transfer Network},
  author={Fang, Xiang and Fang, Wanlong and Wang, Changshuo and Liu, Daizong and Tang, Keke and Dong, Jianfeng and Zhou, Pan and Li, Beibei},
  booktitle={Proceedings of the AAAI Conference on Artificial Intelligence},
  year={2025}
}

@inproceedings{fang2025turing,
  title={Turing Patterns for Multimedia: Reaction-Diffusion Multi-Modal Fusion for Language-Guided Video Moment Retrieval},
  author={Fang, Xiang and Fang, Wanlong and Ji, Wei and Chua, Tat-Seng},
  booktitle={ACM International Conference on Multimedia},
  year={2025}
}

@inproceedings{fang2024not,
  title={Not all inputs are valid: Towards open-set video moment retrieval using language},
  author={Fang, Xiang and Fang, Wanlong and Liu, Daizong and Qu, Xiaoye and Dong, Jianfeng and Zhou, Pan and Li, Renfu and Xu, Zichuan and Chen, Lixing and Zheng, Panpan and others},
  booktitle={Proceedings of the 32nd ACM International Conference on Multimedia},
  pages={28--37},
  year={2024}
}

@inproceedings{fang2023annotations,
  title={Annotations Are Not All You Need: A Cross-modal Knowledge Transfer Network for Unsupervised Temporal Sentence Grounding},
  author={Fang, Xiang and Liu, Daizong and Fang, Wanlong and Zhou, Pan and Cheng, Yu and Tang, Keke and Zou, Kai},
  booktitle={Findings of the Association for Computational Linguistics: EMNLP 2023},
  pages={8721--8733},
  year={2023}
}

@article{fang2021unbalanced,
  title={Unbalanced incomplete multi-view clustering via the scheme of view evolution: Weak views are meat; strong views do eat},
  author={Fang, Xiang and Hu, Yuchong and Zhou, Pan and Wu, Dapeng Oliver},
  journal={IEEE Transactions on Emerging Topics in Computational Intelligence},
  volume={6},
  number={4},
  pages={913--927},
  year={2021},
  publisher={IEEE}
}

@article{fang2025adaptivetai,
  title={Adaptive Hierarchical Graph Cut for Multi-granularity Out-of-distribution Detection},
  author={Fang, Xiang and Easwaran, Arvind and Genest, Blaise and Suganthan, Ponnuthurai Nagaratnam},
  journal={IEEE Transactions on Artificial Intelligence},
  year={2025}
}

@article{fang2021animc,
  title={Animc: A soft approach for autoweighted noisy and incomplete multiview clustering},
  author={Fang, Xiang and Hu, Yuchong and Zhou, Pan and Wu, Dapeng},
  journal={IEEE Transactions on Artificial Intelligence},
  volume={3},
  number={2},
  pages={192--206},
  year={2021},
  publisher={IEEE}
}

@article{fang2020v,
  title={V3H: View variation and view heredity for incomplete multiview clustering},
  author={Fang, Xiang and Hu, Yuchong and Zhou, Pan and Wu, Dapeng Oliver},
  journal={IEEE Transactions on Artificial Intelligence},
  volume={1},
  number={3},
  pages={233--247},
  year={2020},
  publisher={IEEE}
}

@article{fang2020double,
  title={Double self-weighted multi-view clustering via adaptive view fusion},
  author={Fang, Xiang and Hu, Yuchong},
  journal={arXiv preprint arXiv:2011.10396},
  year={2020}
}

@article{liu2023exploring,
  title={Exploring optical-flow-guided motion and detection-based appearance for temporal sentence grounding},
  author={Liu, Daizong and Fang, Xiang and Hu, Wei and Zhou, Pan},
  journal={IEEE Transactions on Multimedia},
  volume={25},
  pages={8539--8553},
  year={2023},
  publisher={IEEE}
}

@inproceedings{wang2025taylor,
  title={Taylor series-inspired local structure fitting network for few-shot point cloud semantic segmentation},
  author={Wang, Changshuo and He, Shuting and Fang, Xiang and Wu, Meiqing and Lam, Siew-Kei and Tiwari, Prayag},
  booktitle={Proceedings of the AAAI Conference on Artificial Intelligence},
  volume={39},
  number={7},
  pages={7527--7535},
  year={2025}
}

@inproceedings{wang2025point,
  title={Point clouds meets physics: Dynamic acoustic field fitting network for point cloud understanding},
  author={Wang, Changshuo and He, Shuting and Fang, Xiang and Han, Jiawei and Liu, Zhonghang and Ning, Xin and Li, Weijun and Tiwari, Prayag},
  booktitle={Proceedings of the Computer Vision and Pattern Recognition Conference},
  pages={22182--22192},
  year={2025}
}

@inproceedings{wang2025dypolyseg,
  title={DyPolySeg: Taylor Series-Inspired Dynamic Polynomial Fitting Network for Few-shot Point Cloud Semantic Segmentation},
  author={Wang, Changshuo and Fang, Xiang and Tiwari, Prayag},
  booktitle={Forty-second International Conference on Machine Learning},
  year={2025}
}

@article{wang2026reasoning,
  title={Reasoning beyond points: A visual introspective approach for few-shot 3d segmentation},
  author={Wang, Changshuo and He, Shuting and Fang, Xiang and Hu, Zhijian and Huang, Jia-Hong and Shen, Yixian and Tiwari, Prayag},
  journal={Advances in Neural Information Processing Systems},
  volume={38},
  pages={117394--117414},
  year={2026}
}

@article{wang2026from,
  title={From Coarse to Fine: Deep Prototype Refinement Network for Few-Shot Point Cloud Semantic Segmentation},
  author={Wang, Changshuo and He, Shuting and Fang, Xiang and Li, Weijun and Gao, Xingyu and Liu, Zhonghang and Tiwari, Prayag and Kanoulas, Dimitrios},
  journal={International Conference on Machine Learning},
  year={2026}
}

@article{wang2026topadapter,
  title={TopAdapter: Topology-Aware Prompt Tuning for Efficient Point Cloud Understanding},
  author={Wang, Changshuo and He, Shuting and Fang, Xiang and Li, Weijun and Shen, Yixian and Xu, Mingkun and Sun, Zhongtian and Tiwari, Prayag},
  journal={International Conference on Machine Learning},
  year={2026}
}

@inproceedings{wang2026biologically,
  title={Biologically-Inspired Evolutionary Domain Symbiosis for Few-shot and Zero-shot Point Cloud Semantic Segmentation},
  author={Wang, Changshuo and Hu, Zhijian and Fang, Xiang and Yu, Zai Yang and Wu, Yibin and Xu, Mingkun and Wang, Yusong and Gao, Xingyu and Tiwari, Prayag},
  booktitle={Proceedings of the AAAI Conference on Artificial Intelligence},
  volume={40},
  number={12},
  pages={9666--9674},
  year={2026}
}

@inproceedings{yang2025eood,
  title={EOOD: Entropy-based Out-of-distribution Detection},
  author={Yang, Guide and Hou, Chao and Peng, Weilong and Fang, Xiang and Nie, Yongwei and Zhu, Peican and Tang, Keke},
  booktitle={2025 International Joint Conference on Neural Networks (IJCNN)},
  pages={1--8},
  year={2025},
  organization={IEEE}
}

@inproceedings{wang2025reducing
,
  title={Reducing T-Depth and T-Count in Quantum Multiplication Using Compressor Primitives},
  author={Wang, Siyi and Dutta, Suman and Lee, Wei Jie Bryan and Feng, Jerrie and Fang, Xiang and Chattopadhyay, Anupam},
  booktitle={Proceedings of the Great Lakes Symposium on VLSI 2025},
  pages={35--40},
  year={2025}
}

@inproceedings{lei2025exploring,
  title={Exploring Disentangled Appearance-Motion Contexts for Temporal Activity Localization},
  author={Lei, Huashuo and Cai, Xiaowen and Liu, Daizong and Fang, Xiang and Qu, Xiaoye and Dong, Jianfeng and Yu, Jixiang and Jin, Keyan},
  booktitle={2025 International Joint Conference on Neural Networks (IJCNN)},
  pages={1--8},
  year={2025},
  organization={IEEE}
}

@inproceedings{zhang2025monoattack,
  title={MonoAttack: A Strong Attack Framework with Depth-Migration and Attribute-Tampering for Monocular 3D Object Detection},
  author={Zhang, Xiayue and Lei, Huashuo and Liu, Daizong and Qu, Xiaoye and Fang, Xiang and Guan, Runwei and Jin, Keyan},
  booktitle={2025 International Joint Conference on Neural Networks (IJCNN)},
  pages={1--8},
  year={2025},
  organization={IEEE}
}

@inproceedings{zhang2025manipulating,
  title={Manipulating the Bounding Box: Multimodal Controlled Backdoor Attacks on 3D Visual Grounding Models},
  author={Zhang, Xiayue and Lei, Huashuo and Liu, Daizong and Qu, Xiaoye and Fang, Xiang and Guan, Runwei and Jin, Keyan},
  booktitle={2025 International Joint Conference on Neural Networks (IJCNN)},
  pages={1--8},
  year={2025},
  organization={IEEE}
}

@article{wang2025prototype,
  title={Prototype-driven structure synergy network for remote sensing images segmentation},
  author={Wang, Junyi and Li, Jinjiang and Fan, Guodong and Ju, Yakun and Fang, Xiang and Kot, Alex C},
  journal={IEEE Transactions on Geoscience and Remote Sensing},
  year={2025},
  publisher={IEEE}
}

@inproceedings{wang2025seeing,
  title={Seeing the Overlooked: Bio-Visual Inspired Weak Saliency Feedback Transformer for Person Re-identification},
  author={Wang, Changshuo and He, Shuting and Fang, Xiang and Nan, Fangzhe and Tiwari, Prayag},
  booktitle={Proceedings of the 33rd ACM International Conference on Multimedia},
  pages={3192--3201},
  year={2025}
}

@inproceedings{fang2026align,
  title={To align or not to align: Strategic multimodal representation alignment for optimal performance},
  author={Fang, Wanlong and Zhang, Tianle and Chan, Alvin},
  booktitle={Proceedings of the AAAI Conference on Artificial Intelligence},
  volume={40},
  number={25},
  pages={21056--21064},
  year={2026}
}

@article{liu2023conditional,
  title={Conditional video diffusion network for fine-grained temporal sentence grounding},
  author={Liu, Daizong and Zhu, Jiahao and Fang, Xiang and Xiong, Zeyu and Wang, Huan and Li, Renfu and Zhou, Pan},
  journal={IEEE Transactions on Multimedia},
  volume={26},
  pages={5461--5476},
  year={2023},
  publisher={IEEE}
}

@article{liu2024pandora,
  title={Pandora's box: Towards building universal attackers against real-world large vision-language models},
  author={Liu, Daizong and Yang, Mingyu and Qu, Xiaoye and Zhou, Pan and Fang, Xiang and Tang, Keke and Wan, Yao and Sun, Lichao},
  journal={Advances in Neural Information Processing Systems},
  volume={37},
  pages={52127--52158},
  year={2024}
}

@inproceedings{liu2026attacking,
  title={Attacking Gray-Box Large Vision-Language Models with Adaptive SVD-Structured Adversarial Alignment},
  author={Liu, Daizong and Cai, Xiaowen and Dong, Junhao and Guo, Zhongliang and Qu, Xiaoye and Guan, Runwei and Fang, Xiang and Ye, Dengpan},
  booktitle={International Conference on Machine Learning},
  year={2026}
}

@inproceedings{liu2024unsupervised,
  title={Unsupervised domain adaptative temporal sentence localization with mutual information maximization},
  author={Liu, Daizong and Fang, Xiang and Qu, Xiaoye and Dong, Jianfeng and Yan, He and Yang, Yang and Zhou, Pan and Cheng, Yu},
  booktitle={Proceedings of the AAAI Conference on Artificial Intelligence},
  volume={38},
  number={4},
  pages={3567--3575},
  year={2024}
}

@inproceedings{liu2023hypotheses,
  title={Hypotheses tree building for one-shot temporal sentence localization},
  author={Liu, Daizong and Fang, Xiang and Zhou, Pan and Di, Xing and Lu, Weining and Cheng, Yu},
  booktitle={Proceedings of the AAAI Conference on Artificial Intelligence},
  volume={37},
  number={2},
  pages={1640--1648},
  year={2023}
}

@inproceedings{tang2024reparameterization,
  title={Reparameterization head for efficient multi-input networks},
  author={Tang, Keke and Zhao, Wenyu and Peng, Weilong and Fang, Xiang and Cui, Xiaodong and Zhu, Peican and Tian, Zhihong},
  booktitle={ICASSP 2024-2024 IEEE International Conference on Acoustics, Speech and Signal Processing (ICASSP)},
  pages={6190--6194},
  year={2024},
  organization={IEEE}
}

@article{xiong2024rethinking,
  title={Rethinking video sentence grounding from a tracking perspective with memory network and masked attention},
  author={Xiong, Zeyu and Liu, Daizong and Fang, Xiang and Qu, Xiaoye and Dong, Jianfeng and Zhu, Jiahao and Tang, Keke and Zhou, Pan},
  journal={IEEE Transactions on Multimedia},
  volume={26},
  pages={11204--11218},
  year={2024},
  publisher={IEEE}
}

@inproceedings{tang2025simplification,
  title={Simplification is all you need against out-of-distribution overconfidence},
  author={Tang, Keke and Hou, Chao and Peng, Weilong and Fang, Xiang and Wu, Zhize and Nie, Yongwei and Wang, Wenping and Tian, Zhihong},
  booktitle={Proceedings of the Computer Vision and Pattern Recognition Conference},
  pages={5030--5040},
  year={2025}
}

@article{cai2026towards,
  title={Towards building model/prompt-transferable attackers against large vision-language models},
  author={Cai, Xiaowen and Liu, Daizong and Qu, Xiaoye and Fang, Xiang and Dong, Jianfeng and Tang, Keke and Zhou, Pan and Sun, Lichao and Hu, Wei},
  journal={Advances in Neural Information Processing Systems},
  volume={38},
  pages={174022--174058},
  year={2026}
}

@article{yan2026fit,
  title={Fit the distribution: Cross-image/prompt adversarial attacks on multimodal large language models},
  author={Yan, Hai and Ma, Haijian and Cai, Xiaowen and Liu, Daizong and Yuan, Zenghui and Qu, Xiaoye and Dong, Jianfeng and Guan, Runwei and Fang, Xiang and He, Hongyang and others},
  journal={Advances in Neural Information Processing Systems},
  volume={38},
  pages={75204--75247},
  year={2026}
}

@inproceedings{liu2024towards,
  title={Towards robust temporal activity localization learning with noisy labels},
  author={Liu, Daizong and Qu, Xiaoye and Fang, Xiang and Dong, Jianfeng and Zhou, Pan and Nan, Guoshun and Tang, Keke and Fang, Wanlong and Cheng, Yu},
  booktitle={Proceedings of the 2024 Joint International Conference on Computational Linguistics, Language Resources and Evaluation (LREC-COLING 2024)},
  pages={16630--16642},
  year={2024}
}

@inproceedings{cai2025imperceptible,
  title={Imperceptible Beam-Sensitive Adversarial Attacks for LiDAR-based Object Detection in Autonomous Driving},
  author={Cai, Fuyao and Liu, Daizong and Fang, Xiang and Yu, Jixiang and Tang, Keke and Zhou, Pan},
  booktitle={2025 IEEE International Conference on Multimedia and Expo (ICME)},
  pages={1--6},
  year={2025},
  organization={IEEE}
}

@article{kuai2026dynamic,
  title={Dynamic Graph-enhanced Event Refinement for Temporal Sentence Grounding of Micro-moments},
  author={Kuai, Mingjin and Qin, You and Fang, Xiang and Ji, Wei and Zimmermann, Roger},
  journal={IEEE Transactions on Multimedia},
  year={2026},
  publisher={IEEE}
}

@inproceedings{fang2026towardsicml,
  title={Towards Understanding Modality Interaction in Multimodal Language Models via Partial Information Decomposition},
  author={Fang, Wanlong and Zhang, Tianle and Tao, Wen and Chan, Alvin},
  booktitle={International Conference on Machine Learning},
  year={2026}
}

@article{dong2018predicting,
  title={Predicting visual features from text for image and video caption retrieval},
  author={Dong, Jianfeng and Li, Xirong and Snoek, Cees GM},
  journal={IEEE Transactions on Multimedia},
  volume={20},
  number={12},
  pages={3377--3388},
  year={2018},
  publisher={IEEE}
}

@article{dong2022reading,
  title={Reading-strategy inspired visual representation learning for text-to-video retrieval},
  author={Dong, Jianfeng and Wang, Yabing and Chen, Xianke and Qu, Xiaoye and Li, Xirong and He, Yuan and Wang, Xun},
  journal={IEEE Transactions on Circuits and Systems for Video Technology},
  volume={32},
  number={8},
  pages={5680--5694},
  year={2022}
}

@article{zheng2023progressive,
  title={Progressive localization networks for language-based moment localization},
  author={Zheng, Qi and Dong, Jianfeng and Qu, Xiaoye and Yang, Xun and Wang, Yabing and Zhou, Pan and Liu, Baolong and Wang, Xun},
  journal={ACM Transactions on Multimedia Computing, Communications and Applications},
  volume={19},
  number={2},
  pages={1--21},
  year={2023}
}

@inproceedings{liu2021context,
  title={Context-aware biaffine localizing network for temporal sentence grounding},
  author={Liu, Daizong and Qu, Xiaoye and Dong, Jianfeng and Zhou, Pan and Cheng, Yu and Wei, Wei and Xu, Zichuan and Xie, Yulai},
  booktitle={Proceedings of the IEEE/CVF Conference on Computer Vision and Pattern Recognition},
  pages={11235--11244},
  year={2021}
}

@inproceedings{dong2022partially,
  title={Partially Relevant Video Retrieval},
  author={Dong, Jianfeng and Chen, Xianke and Zhang, Minsong and Yang, Xun and Chen, Shujie and Li, Xirong and Wang, Xun},
  booktitle={Proceedings of the 30th ACM International Conference on Multimedia},
  pages={246--257},
  year={2022}
}

@inproceedings{dong2023hierarchical,
  title={Hierarchical contrast for unsupervised skeleton-based action representation learning},
  author={Dong, Jianfeng and Sun, Shengkai and Liu, Zhonglin and Chen, Shujie and Liu, Baolong and Wang, Xun},
  booktitle={Proceedings of the AAAI Conference on Artificial Intelligence},
  volume={37},
  number={1},
  pages={525--533},
  year={2023}
}

@inproceedings{dong2023region,
  title={From Region to Patch: Attribute-Aware Foreground-Background Contrastive Learning for Fine-Grained Fashion Retrieval},
  author={Dong, Jianfeng and Peng, Xiaoman and Ma, Zhe and Liu, Daizong and Qu, Xiaoye and Yang, Xun and Zhu, Jixiang and Liu, Baolong},
  booktitle={Proceedings of the 46th International ACM SIGIR Conference on Research and Development in Information Retrieval},
  pages={1273--1282},
  year={2023}
}

@article{wang2022learning,
  title={Learning discriminative features by covering local geometric space for point cloud analysis},
  author={Wang, Changshuo and Ning, Xin and Sun, Linjun and Zhang, Liping and Li, Weijun and Bai, Xiao},
  journal={IEEE Transactions on Geoscience and Remote Sensing},
  volume={60},
  pages={1--15},
  year={2022},
  publisher={IEEE}
}

@article{wangchangshuo20223d,
  title={3d point cloud classification method based on dynamic coverage of local area},
  author={Wang, Changshuo and Wang, Han and Ning, Xin and Tian Shengwei and Li, Weijun},
  journal={Journal of Software},
  volume={34},
  number={4},
  pages={1962--1976},
  year={2022}
}

@article{wang20233d,
  title={3D person re-identification based on global semantic guidance and local feature aggregation},
  author={Wang, Changshuo and Ning, Xin and Li, Weijun and Bai, Xiao and Gao, Xingyu},
  journal={IEEE Transactions on Circuits and Systems for Video Technology},
  year={2023},
  publisher={IEEE}
}

@article{wang2021brief,
  title={A brief survey on RGB-D semantic segmentation using deep learning},
  author={Wang, Changshuo and Wang, Chen and Li, Weijun and Wang, Haining},
  journal={Displays},
  volume={70},
  pages={102080},
  year={2021},
  publisher={Elsevier}
}

@inproceedings{jiang2023lttpoint,
  title={LTTPoint: A MLP-Based Point Cloud Classification Method with Local Topology Transformation Module},
  author={Jiang, Limin and Wang, Changshuo and Ning, Xin and Yu, Zaiyang},
  booktitle={2023 7th Asian Conference on Artificial Intelligence Technology (ACAIT)},
  pages={783--789},
  year={2023},
  organization={IEEE}
}

@article{zhang2024pointgt,
  title={PointGT: A Method for Point-Cloud Classification and Segmentation Based on Local Geometric Transformation},
  author={Zhang, Huang and Wang, Changshuo and Yu, Long and Tian, Shengwei and Ning, Xin and Rodrigues, Joel},
  journal={IEEE Transactions on Multimedia},
  year={2024},
  publisher={IEEE}
}

@article{zhang2023deep,
  title={Deep learning-based 3D point cloud classification: A systematic survey and outlook},
  author={Zhang, Huang and Wang, Changshuo and Tian, Shengwei and Lu, Baoli and Zhang, Liping and Ning, Xin and Bai, Xiao},
  journal={Displays},
  volume={79},
  pages={102456},
  year={2023},
  publisher={Elsevier}
}

@article{ning2023occluded,
  title={Occluded person re-identification with deep learning: a survey and perspectives},
  author={Ning, Enhao and Wang, Changshuo and Zhang, Huang and Ning, Xin and Tiwari, Prayag},
  journal={Expert Systems with Applications},
  pages={122419},
  year={2023},
  publisher={Elsevier}
}

@article{ning2024enhancement,
  title={Enhancement, integration, expansion: Activating representation of detailed features for occluded person re-identification},
  author={Ning, Enhao and Wang, Yangfan and Wang, Changshuo and Zhang, Huang and Ning, Xin},
  journal={Neural Networks},
  volume={169},
  pages={532--541},
  year={2024},
  publisher={Elsevier}
}

@article{ning2023pedestrian,
  title={Pedestrian Re-ID based on feature consistency and contrast enhancement},
  author={Ning, Enhao and Zhang, Canlong and Wang, Changshuo and Ning, Xin and Chen, Hao and Bai, Xiao},
  journal={Displays},
  volume={79},
  pages={102467},
  year={2023},
  publisher={Elsevier}
}

@article{yu2024pedestrian,
  title={Pedestrian 3d shape understanding for person re-identification via multi-view learning},
  author={Yu, Zaiyang and Li, Lusi and Xie, Jinlong and Wang, Changshuo and Li, Weijun and Ning, Xin},
  journal={IEEE Transactions on Circuits and Systems for Video Technology},
  year={2024},
  publisher={IEEE}
}

@inproceedings{liu2022learning,
  title={Learning to focus on the foreground for temporal sentence grounding},
  author={Liu, Daizong and Hu, Wei},
  booktitle={Proceedings of the 29th International Conference on Computational Linguistics},
  pages={5532--5541},
  year={2022}
}

@inproceedings{guo2022hybird,
  title={A hybird alignment loss for temporal moment localization with natural language},
  author={Guo, Chao and Liu, Daizong and Zhou, Pan},
  booktitle={2022 IEEE International Conference on Multimedia and Expo (ICME)},
  pages={1--6},
  year={2022},
  organization={IEEE}
}

@inproceedings{xiong2023tracking,
  title={Tracking Objects and Activities with Attention for Temporal Sentence Grounding},
  author={Xiong, Zeyu and Liu, Daizong and Zhou, Pan and Zhu, Jiahao},
  booktitle={ICASSP 2023-2023 IEEE International Conference on Acoustics, Speech and Signal Processing (ICASSP)},
  pages={1--5},
  year={2023},
  organization={IEEE}
}

@inproceedings{liu2023jointly,
  title={Jointly visual-and semantic-aware graph memory networks for temporal sentence localization in videos},
  author={Liu, Daizong and Zhou, Pan},
  booktitle={ICASSP 2023-2023 IEEE International Conference on Acoustics, Speech and Signal Processing (ICASSP)},
  pages={1--5},
  year={2023},
  organization={IEEE}
}

@ARTICLE{10042189,
  author={Deng, Shijie and Wen, Jie and Liu, Chengliang and Yan, Ke and Xu, Gehui and Xu, Yong},
  journal={IEEE Transactions on Neural Networks and Learning Systems}, 
  title={Projective Incomplete Multi-View Clustering}, 
  year={2023},
  volume={},
  number={},
  pages={1-13},
  doi={10.1109/TNNLS.2023.3242473}}

@inproceedings{liu2023incomplete,
  title={Incomplete Multi-View Multi-Label Learning via Label-Guided Masked View- and Category-Aware Transformers},
  author={Liu, Chengliang and Wen, Jie and Luo, Xiaoling and Xu, Yong},
  booktitle={Proceedings of the AAAI Conference on Artificial Intelligence},
  year={2023},
  volume={37},
  number={7},
  pages={8816--8824}
}

@article{tang2022decision,
  title={Decision fusion networks for image classification},
  author={Tang, Keke and Ma, Yuexin and Miao, Dingruibo and Song, Peng and Gu, Zhaoquan and Tian, Zhihong and Wang, Wenping},
  journal={IEEE Transactions on Neural Networks and Learning Systems},
 doi={10.1109/TNNLS.2022.3196129},
  year={2022}
}

@article{tang2024effective,
  title={Effective Single-Step Adversarial Training With Energy-Based Models},
  author={Tang, Keke and Lou, Tianrui and Peng, Weilong and Chen, Nenglun and Shi, Yawen and Wang, Wenping},
  journal={IEEE Transactions on Emerging Topics in Computational Intelligence},
doi={10.1109/TETCI.2024.3378652},
  year={2024}
}

@article{tang2022rethinking,
  title={Rethinking perturbation directions for imperceptible adversarial attacks on point clouds},
  author={Tang, Keke and Shi, Yawen and Lou, Tianrui and Peng, Weilong and He, Xu and Zhu, Peican and Gu, Zhaoquan and Tian, Zhihong},
  journal={IEEE Internet of Things Journal},
  volume={10},
  number={6},
  pages={5158--5169},
  year={2022}
}

@article{tang2023reppvconv,
  title={Reppvconv: attentively fusing reparameterized voxel features for efficient 3d point cloud perception},
  author={Tang, Keke and Chen, Yuhong and Peng, Weilong and Zhang, Yanling and Fang, Meie and Wang, Zheng and Song, Peng},
  journal={The Visual Computer},
  volume={39},
  number={11},
  pages={5577--5588},
  year={2023}
}

@inproceedings{liu2023dicnet,
  title={DICNet: Deep Instance-Level Contrastive Network for Double Incomplete Multi-View Multi-Label Classification},
  author={Liu, Chengliang and Wen, Jie and Luo, Xiaoling and Huang, Chao and Wu, Zhihao and Xu, Yong},
  booktitle={Proceedings of the AAAI Conference on Artificial Intelligence},
  volume={37},
  number={7},
  pages={8807--8815},
  year={2023}
}

@article{liu2023information,
  author={Liu, Chengliang and Wen, Jie and Wu, Zhihao and Luo, Xiaoling and Huang, Chao and Xu, Yong},
  journal={IEEE Transactions on Neural Networks and Learning Systems}, 
  title={Information Recovery-Driven Deep Incomplete Multiview Clustering Network}, 
  year={2023},
  volume={},
  number={},
  pages={1-11},
}

@ARTICLE{10086538,
  author={Wen, Jie and Liu, Chengliang and Deng, Shijie and Liu, Yicheng and Fei, Lunke and Yan, Ke and Xu, Yong},
  journal={IEEE Transactions on Neural Networks and Learning Systems}, 
  title={Deep Double Incomplete Multi-View Multi-Label Learning With Incomplete Labels and Missing Views}, 
  year={2023},
  volume={},
  number={},
  pages={1-13},
  doi={10.1109/TNNLS.2023.3260349}}

@article{2023A,
  title={A Survey on Incomplete Multiview Clustering},
  author={ Wen, Jie  and  Zhang, Zheng  and  Li, Zhang Jinxing },
  journal={IEEE transactions on systems, man, and cybernetics. Systems},
  volume={53},
  number={2 Pt.2},
  pages={1136-1149},
  year={2023},
}

@INPROCEEDINGS{10447574,
  author={Tang, Keke and Zhao, Wenyu and Peng, Weilong and Fang, Xiang and Cui, Xiaodong and Zhu, Peican and Tian, Zhihong},
  booktitle={ICASSP 2024 - 2024 IEEE International Conference on Acoustics, Speech and Signal Processing (ICASSP)}, 
  title={Reparameterization Head for Efficient Multi-Input Networks}, 
  year={2024},
  volume={},
  number={},
  pages={6190-6194},
  doi={10.1109/ICASSP48485.2024.10447574}}

@inproceedings{liu2022skimming,
  title={Skimming, locating, then perusing: A human-like framework for natural language video localization},
  author={Liu, Daizong and Hu, Wei},
  booktitle={Proceedings of the 30th ACM International Conference on Multimedia},
  pages={4536--4545},
  year={2022}
}

@inproceedings{liu2022reducing,
  title={Reducing the vision and language bias for temporal sentence grounding},
  author={Liu, Daizong and Qu, Xiaoye and Hu, Wei},
  booktitle={Proceedings of the 30th ACM International Conference on Multimedia},
  pages={4092--4101},
  year={2022}
}

@article{zhu2023rethinking,
  title={Rethinking the video sampling and reasoning strategies for temporal sentence grounding},
  author={Zhu, Jiahao and Liu, Daizong and Zhou, Pan and Di, Xing and Cheng, Yu and Yang, Song and Xu, Wenzheng and Xu, Zichuan and Wan, Yao and Sun, Lichao and others},
  journal={arXiv preprint arXiv:2301.00514},
  year={2023}
}

@article{liu2022few,
  title={Few-shot temporal sentence grounding via memory-guided semantic learning},
  author={Liu, Daizong and Zhou, Pan and Xu, Zichuan and Wang, Haozhao and Li, Ruixuan},
  journal={IEEE Transactions on Circuits and Systems for Video Technology},
  volume={33},
  number={5},
  pages={2491--2505},
  year={2022},
  publisher={IEEE}
}

@inproceedings{liu2023filling,
  title={Filling the Information Gap between Video and Query for Language-Driven Moment Retrieval},
  author={Liu, Daizong and Qu, Xiaoye and Dong, Jianfeng and Nan, Guoshun and Zhou, Pan and Xu, Zichuan and Chen, Lixing and Yan, He and Cheng, Yu},
  booktitle={Proceedings of the 31st ACM International Conference on Multimedia},
  pages={4190--4199},
  year={2023}
}

@article{liu2024transform,
  title={Transform-Equivariant Consistency Learning for Temporal Sentence Grounding},
  author={Liu, Daizong and Qu, Xiaoye and Dong, Jianfeng and Zhou, Pan and Xu, Zichuan and Wang, Haozhao and Di, Xing and Lu, Weining and Cheng, Yu},
  journal={ACM Transactions on Multimedia Computing, Communications and Applications},
  volume={20},
  number={4},
  pages={1--19},
  year={2024},
  publisher={ACM New York, NY}
}

@inproceedings{albarelli2010game,
  title={A game-theoretic approach to fine surface registration without initial motion estimation},
  author={Albarelli, Andrea and Rodola, Emanuele and Torsello, Andrea},
  booktitle={2010 IEEE computer society conference on computer vision and pattern recognition},
  pages={430--437},
  year={2010},
  organization={IEEE}
}

@inproceedings{torsello2006grouping,
  title={Grouping with asymmetric affinities: A game-theoretic perspective},
  author={Torsello, Andrea and Bulo, S Rota and Pelillo, Marcello},
  booktitle={2006 IEEE Computer Society Conference on Computer Vision and Pattern Recognition (CVPR'06)},
  volume={1},
  pages={292--299},
  year={2006},
  organization={IEEE}
}

@inproceedings{lin2023univtg,
  title={UniVTG: Towards Unified Video-Language Temporal Grounding},
  author={Lin, Kevin Qinghong and Zhang, Pengchuan and Chen, Joya and Pramanick, Shraman and Gao, Difei and Wang, Alex Jinpeng and Yan, Rui and Shou, Mike Zheng},
  booktitle={Proceedings of the IEEE/CVF International Conference on Computer Vision},
  pages={2794--2804},
  year={2023}
}

@article{nowak1997axiomatization,
  title={On an axiomatization of the Banzhaf value without the additivity axiom},
  author={Nowak, Andrzej S},
  journal={International Journal of Game Theory},
  volume={26},
  pages={137--141},
  year={1997},
  publisher={Springer}
}

@article{lehrer1988axiomatization,
  title={An axiomatization of the Banzhaf value},
  author={Lehrer, Ehud},
  journal={International Journal of Game Theory},
  volume={17},
  pages={89--99},
  year={1988},
  publisher={Springer}
}

@article{michalak2013efficient,
  title={Efficient computation of the Shapley value for game-theoretic network centrality},
  author={Michalak, Tomasz P and Aadithya, Karthik V and Szczepanski, Piotr L and Ravindran, Balaraman and Jennings, Nicholas R},
  journal={Journal of Artificial Intelligence Research},
  volume={46},
  pages={607--650},
  year={2013}
}

@article{winter2002shapley,
  title={The shapley value},
  author={Winter, Eyal},
  journal={Handbook of game theory with economic applications},
  volume={3},
  pages={2025--2054},
  year={2002},
  publisher={Elsevier}
}

@inproceedings{donoser2013diffusion,
  title={Diffusion processes for retrieval revisited},
  author={Donoser, Michael and Bischof, Horst},
  booktitle={Proceedings of the IEEE conference on computer vision and pattern recognition},
  pages={1320--1327},
  year={2013}
}

@inproceedings{rodola2012game,
  title={A game-theoretic approach to deformable shape matching},
  author={Rodola, Emanuele and Bronstein, Alex M and Albarelli, Andrea and Bergamasco, Filippo and Torsello, Andrea},
  booktitle={2012 IEEE Conference on Computer Vision and Pattern Recognition},
  pages={182--189},
  year={2012},
  organization={IEEE}
}

@inproceedings{dowdall2006coalitional,
  title={Coalitional tracking in facial infrared imaging and beyond},
  author={Dowdall, Jonathan and Pavlidis, Ioannis T and Tsiamyrtzis, Panagiotis},
  booktitle={2006 Conference on Computer Vision and Pattern Recognition Workshop},
  pages={134--134},
  year={2006},
  organization={IEEE}
}

@inproceedings{pavan2003new,
  title={A new graph-theoretic approach to clustering and segmentation},
  author={Pavan, Massimiliano and Pelillo, Marcello},
  booktitle={2003 IEEE Computer Society Conference on Computer Vision and Pattern Recognition, 2003. Proceedings.},
  volume={1},
  pages={I--I},
  year={2003},
  organization={IEEE}
}

@inproceedings{ma2017forecasting,
  title={Forecasting interactive dynamics of pedestrians with fictitious play},
  author={Ma, Wei-Chiu and Huang, De-An and Lee, Namhoon and Kitani, Kris M},
  booktitle={Proceedings of the IEEE Conference on Computer Vision and Pattern Recognition},
  pages={774--782},
  year={2017}
}

@article{jin2021game,
  title={Game theoretical analysis on capacity configuration for microgrid based on multi-agent system},
  author={Jin, Shunping and Wang, Shoupeng and Fang, Fang},
  journal={International Journal of Electrical Power \& Energy Systems},
  volume={125},
  pages={106485},
  year={2021},
  publisher={Elsevier}
}

@inproceedings{patel2021game,
  title={Game-theoretic vocabulary selection via the shapley value and banzhaf index},
  author={Patel, Roma and Garnelo, Marta and Gemp, Ian and Dyer, Chris and Bachrach, Yoram},
  booktitle={Proceedings of the 2021 Conference of the North American Chapter of the Association for Computational Linguistics: Human Language Technologies},
  pages={2789--2798},
  year={2021}
}

@inproceedings{jin2023video,
  title={Video-text as game players: Hierarchical banzhaf interaction for cross-modal representation learning},
  author={Jin, Peng and Huang, Jinfa and Xiong, Pengfei and Tian, Shangxuan and Liu, Chang and Ji, Xiangyang and Yuan, Li and Chen, Jie},
  booktitle={Proceedings of the IEEE/CVF Conference on Computer Vision and Pattern Recognition},
  pages={2472--2482},
  year={2023}
}

@inproceedings{tan2021logan,
  title={Logan: Latent graph co-attention network for weakly-supervised video moment retrieval},
  author={Tan, Reuben and Xu, Huijuan and Saenko, Kate and Plummer, Bryan A},
  booktitle={Proceedings of the IEEE Winter Conference on Applications of Computer Vision},
  pages={2083--2092},
  year={2021}
}

@article{song2020weakly,
  title={Weakly-supervised multi-level attentional reconstruction network for grounding textual queries in videos},
  author={Song, Yijun and Wang, Jingwen and Ma, Lin and Yu, Zhou and Yu, Jun},
  journal={arXiv preprint arXiv:2003.07048},
  year={2020}
}

@article{zhang2020counterfactual,
  title={Counterfactual Contrastive Learning for Weakly-Supervised Vision-Language Grounding},
  author={Zhang, Zhu and Zhao, Zhou and Lin, Zhijie and He, Xiuqiang and others},
  journal={Advances in Neural Information Processing Systems},
  volume={33},
  pages={18123--18134},
  year={2020}
}

@inproceedings{ma2020vlanet,
  title={{VLANet}: Video-language alignment network for weakly-supervised video moment retrieval},
  author={Ma, Minuk and Yoon, Sunjae and Kim, Junyeong and Lee, Youngjoon and Kang, Sunghun and Yoo, Chang D},
  booktitle={Proceedings of the European Conference on Computer Vision},
  pages={156--171},
  year={2020}
}

@article{chen2020look,
  title={Look closer to ground better: Weakly-supervised temporal grounding of sentence in video},
  author={Chen, Zhenfang and Ma, Lin and Luo, Wenhan and Tang, Peng and Wong, Kwan-Yee K},
  journal={arXiv preprint arXiv:2001.09308},
  year={2020}
}

@inproceedings{lin2020weakly,
  title={Weakly-supervised video moment retrieval via semantic completion network},
  author={Lin, Zhijie and Zhao, Zhou and Zhang, Zhu and Wang, Qi and Liu, Huasheng},
  booktitle={Proceedings of the AAAI Conference on Artificial Intelligence},
  volume={34},
  number={07},
  pages={11539--11546},
  year={2020}
}

@inproceedings{gao2019wslln,
  title={WSLLN: Weakly Supervised Natural Language Localization Networks},
  author={Gao, Mingfei and Davis, Larry and Socher, Richard and Xiong, Caiming},
  booktitle={Proceedings of the 2019 Conference on Empirical Methods in Natural Language Processing and the 9th International Joint Conference on Natural Language Processing},
  pages={1481--1487},
  year={2019}
}

@inproceedings{Alpher33,
  author = {Zhang, Songyang and Peng, Houwen and Fu, Jianlong and Luo, Jiebo},
  title = {Learning 2D Temporal Adjacent Networks for Moment Localization with Natural Language.},
  booktitle = {Proceedings of the AAAI Conference on Artificial Intelligence},
  pages = {12870-12877},
  volume={34},
  number={07},
  year = 2020
}

@inproceedings{Alpher08,
  title={Localizing natural language in videos},
  author={Chen, Jingyuan and Ma, Lin and Chen, Xinpeng and Jie, Zequn and Luo, Jiebo},
  booktitle={Proceedings of the AAAI Conference on Artificial Intelligence},
  volume={33},
  number={01},
  pages={8175--8182},
  year={2019}
}

@inproceedings{yuan2020semantic,
  title={Semantic conditioned dynamic modulation for temporal sentence grounding in videos},
  author={Yuan, Yitian and Ma, Lin and Wang, Jingwen and Liu, Wei and Zhu, Wenwu},
  booktitle={Proceedings of the 33rd International Conference on Neural Information Processing Systems},
  pages={536--546},
  year={2019}
}

@inproceedings{wang2020temporally,
  title={Temporally grounding language queries in videos by contextual boundary-aware prediction},
  author={Wang, Jingwen and Ma, Lin and Jiang, Wenhao},
  booktitle={Proceedings of the AAAI Conference on Artificial Intelligence},
  volume={34},
  pages={12168--12175},
  year={2020}
}

@article{Alpher07,
  author = {Hendricks, Lisa Anne and Wang, Oliver and Shechtman, Eli and Sivic, Josef and Darrell, Trevor and Russell, Bryan},
  title = {Localizing moments in video with temporal language.},
  journal = {Proceedings of the 2018 Conference on Empirical Methods in Natural Language Processing},
  pages = {1380–1390}, 
  year = 2018
}

@ARTICLE{yang2021local,
  author={Yang, Wenfei and Zhang, Tianzhu and Zhang, Yongdong and Wu, Feng},
  journal={IEEE Transactions on Image Processing},
  title={Local Correspondence Network for Weakly Supervised Temporal Sentence Grounding}, 
  year={2021},
  volume={30},
  number={},
  pages={3252-3262}
 }

@article{wang2021weakly,
  title={Weakly supervised temporal adjacent network for language grounding},
  author={Wang, Yuechen and Deng, Jiajun and Zhou, Wengang and Li, Houqiang},
  journal={IEEE Transactions on Multimedia},
  volume={24},
  pages={3276--3286},
  year={2021},
  publisher={IEEE}
}

@inproceedings{zheng2022cnm,
  title={Weakly supervised video moment localization with contrastive negative sample mining},
  author={Zheng, Minghang and Huang, Yanjie and Chen, Qingchao and Liu, Yang},
  booktitle={Proceedings of the AAAI Conference on Artificial Intelligence},
  volume={36},
  number={3},
  pages={3517--3525},
  year={2022}
}

@inproceedings{zheng2022cpl,
  title={Weakly Supervised Temporal Sentence Grounding with Gaussian-based Contrastive Proposal Learning},
  author={Zheng, Minghang and Huang, Yanjie and Chen, Qingchao and Peng, Yuxin and Liu, Yang},
  booktitle={Proceedings of the IEEE/CVF Conference on Computer Vision and Pattern Recognition},
  pages={15555--15564},
  year={2022}
}

@article{guo2024benchmarking,
  title={Benchmarking Micro-action Recognition: Dataset, Method, and Application},
  author={Guo, Dan and Li, Kun and Hu, Bin and Zhang, Yan and Wang, Meng},
  journal={IEEE Transactions on Circuits and Systems for Video Technology},
  volume={34},
  number={7},
  pages={6238--6252},
  year={2024},
  publisher={IEEE}
}

@article{leech2002computation,
  title={Computation of power indices},
  author={Leech, Dennis},
  year={2002},
  publisher={Warwick Economic Research Papers}
}

@article{bachrach2010approximating,
  title={Approximating power indices: theoretical and empirical analysis},
  author={Bachrach, Yoram and Markakis, Evangelos and Resnick, Ezra and Procaccia, Ariel D and Rosenschein, Jeffrey S and Saberi, Amin},
  journal={Autonomous Agents and Multi-Agent Systems},
  volume={20},
  pages={105--122},
  year={2010},
  publisher={Springer}
}

@article{song2023marn,
  title={MARN: Multi-level Attentional Reconstruction Networks for Weakly Supervised Video Temporal Grounding},
  author={Song, Yijun and Wang, Jingwen and Ma, Lin and Yu, Jun and Liang, Jinxiu and Yuan, Liu and Yu, Zhou},
  journal={Neurocomputing},
  volume={554},
  pages={126625},
  year={2023},
  publisher={Elsevier}
}

@inproceedings{wu2023atomic,
  title={Atomic-action-based Contrastive Network for Weakly Supervised Temporal Language Grounding},
  author={Wu, Hongzhou and Lyu, Yifan and Shen, Xingyu and Zhao, Xuechen and Wang, Mengzhu and Zhang, Xiang and Luo, Zhigang},
  booktitle={2023 IEEE International Conference on Multimedia and Expo (ICME)},
  pages={1523--1528},
  year={2023},
  organization={IEEE}
}

@article{ma2023dual,
  title={Dual Masked Modeling for Weakly-Supervised Temporal Boundary Discovery},
  author={Ma, Yuer and Liu, Yi and Wang, Limin and Kang, Wenxiong and Qiao, Yu and Wang, Yali},
  journal={IEEE Transactions on Multimedia},
  year={2023},
  publisher={IEEE}
}

@inproceedings{li2023g2l,
  title={G2l: Semantically aligned and uniform video grounding via geodesic and game theory},
  author={Li, Hongxiang and Cao, Meng and Cheng, Xuxin and Li, Yaowei and Zhu, Zhihong and Zou, Yuexian},
  booktitle={Proceedings of the IEEE/CVF International Conference on Computer Vision},
  pages={12032--12042},
  year={2023}
}

@article{kingma2014adam,
  title={Adam: A method for stochastic optimization},
  author={Kingma, Diederik P and Ba, Jimmy},
  journal={arXiv preprint arXiv:1412.6980},
  year={2014}
}

@inproceedings{tran2015learning,
  title={Learning spatiotemporal features with 3d convolutional networks},
  author={Tran, Du and Bourdev, Lubomir and Fergus, Rob and Torresani, Lorenzo and Paluri, Manohar},
  booktitle={Proceedings of the IEEE International Conference on Computer Vision},
  pages={4489--4497},
  year={2015}
}

@inproceedings{vaswani2017attention,
  title={Attention is all you need},
  author={Vaswani, Ashish and Shazeer, Noam and Parmar, Niki and Uszkoreit, Jakob and Jones, Llion and Gomez, Aidan N and Kaiser, {\L}ukasz and Polosukhin, Illia},
  booktitle={Advances in Neural Information Processing Systems},
  pages={5998--6008},
  year={2017}
}

@inproceedings{zhang2019cross,
  title={Cross-modal interaction networks for query-based moment retrieval in videos},
  author={Zhang, Zhu and Lin, Zhijie and Zhao, Zhou and Xiao, Zhenxin},
  booktitle={Proceedings of the 42nd International ACM SIGIR Conference on Research and Development in Information Retrieval},
  pages={655--664},
  year={2019}
}

@inproceedings{pennington2014glove,
  title={Glove: Global vectors for word representation},
  author={Pennington, Jeffrey and Socher, Richard and Manning, Christopher D},
  booktitle={Proceedings of the Conference on Empirical Methods in Natural Language Processing},
  pages={1532--1543},
  year={2014}
}

@inproceedings{krishna2017dense,
  title={Dense-captioning events in videos},
  author={Krishna, Ranjay and Hata, Kenji and Ren, Frederic and Fei-Fei, Li and Carlos Niebles, Juan},
  booktitle={Proceedings of the IEEE International Conference on Computer Vision},
  pages={706--715},
  year={2017}
}

@inproceedings{gao2017tall,
  title={Tall: Temporal activity localization via language query},
  author={Gao, Jiyang and Sun, Chen and Yang, Zhenheng and Nevatia, Ram},
  booktitle={Proceedings of the IEEE International Conference on Computer Vision},
  pages={5267--5275},
  year={2017}
}

@inproceedings{anne2017localizing,
  title={Localizing moments in video with natural language},
  author={Anne Hendricks, Lisa and Wang, Oliver and Shechtman, Eli and Sivic, Josef and Darrell, Trevor and Russell, Bryan},
  booktitle={Proceedings of the IEEE International Conference on Computer Vision},
  pages={5803--5812},
  year={2017}
}

@inproceedings{zhang2019man,
  title={Man: Moment alignment network for natural language moment retrieval via iterative graph adjustment},
  author={Zhang, Da and Dai, Xiyang and Wang, Xin and Wang, Yuan-Fang and Davis, Larry S},
  booktitle={Proceedings of the IEEE Conference on Computer Vision and Pattern Recognition},
  pages={1247--1257},
  year={2019}
}

@inproceedings{mithun2019weakly,
  title={Weakly supervised video moment retrieval from text queries},
  author={Mithun, Niluthpol Chowdhury and Paul, Sujoy and Roy-Chowdhury, Amit K},
  booktitle={Proceedings of the IEEE Conference on Computer Vision and Pattern Recognition},
  pages={11592--11601},
  year={2019}
}

@inproceedings{chen2020rethinking,
  title={Rethinking the bottom-up framework for query-based video localization},
  author={Chen, Long and Lu, Chujie and Tang, Siliang and Xiao, Jun and Zhang, Dong and Tan, Chilie and Li, Xiaolin},
  booktitle={Proceedings of the AAAI Conference on Artificial Intelligence},
  volume={34},
  number={07},
  pages={10551--10558},
  year={2020}
}

@inproceedings{sigurdsson2016hollywood,
  title={Hollywood in homes: Crowdsourcing data collection for activity understanding},
  author={Sigurdsson, Gunnar A and Varol, G{\"u}l and Wang, Xiaolong and Farhadi, Ali and Laptev, Ivan and Gupta, Abhinav},
  booktitle={European Conference on Computer Vision},
  pages={510--526},
  year={2016}
}

@inproceedings{carreira2017quo,
  title={Quo vadis, action recognition? a new model and the kinetics dataset},
  author={Carreira, Joao and Zisserman, Andrew},
  booktitle={proceedings of the IEEE Conference on Computer Vision and Pattern Recognition},
  pages={6299--6308},
  year={2017}
}

@inproceedings{mun2020local,
  title={Local-Global Video-Text Interactions for Temporal Grounding},
  author={Mun, Jonghwan and Cho, Minsu and Han, Bohyung},
  booktitle={Proceedings of the IEEE Conference on Computer Vision and Pattern Recognition},
  pages={10810--10819},
  year={2020}
}

@inproceedings{zeng2020dense,
  title={Dense regression network for video grounding},
  author={Zeng, Runhao and Xu, Haoming and Huang, Wenbing and Chen, Peihao and Tan, Mingkui and Gan, Chuang},
  booktitle={Proceedings of the IEEE Conference on Computer Vision and Pattern Recognition},
  pages={10287--10296},
  year={2020}
}

@inproceedings{liu2020jointly,
  title={Jointly Cross-and Self-Modal Graph Attention Network for Query-Based Moment Localization},
  author={Liu, Daizong and Qu, Xiaoye and Liu, Xiao-Yang and Dong, Jianfeng and Zhou, Pan and Xu, Zichuan},
  booktitle={Proceedings of the 28th ACM International Conference on Multimedia},
  pages={4070--4078},
  year={2020}
}

@inproceedings{zhang2020span,
  title={Span-based Localizing Network for Natural Language Video Localization},
  author={Zhang, Hao and Sun, Aixin and Jing, Wei and Zhou, Joey Tianyi},
  booktitle={Proceedings of the 58th Annual Meeting of the Association for Computational Linguistics},
  pages={6543--6554},
  year={2020}
}

@inproceedings{xiong2022gaussian,
  title={Gaussian Kernel-Based Cross Modal Network for Spatio-Temporal Video Grounding},
  author={Xiong, Zeyu and Liu, Daizong and Zhou, Pan},
  booktitle={IEEE International Conference on Image Processing (ICIP) },
  year={2022},
  volume={},
  number={},
  pages={2481-2485}
}

@article{grabisch1999axiomatic,
  title={An axiomatic approach to the concept of interaction among players in cooperative games},
  author={Grabisch, Michel and Roubens, Marc},
  journal={International Journal of game theory},
  volume={28},
  pages={547--565},
  year={1999},
  publisher={Springer}
}

@article{chalkiadakis2011computational,
  title={Computational aspects of cooperative game theory},
  author={Chalkiadakis, Georgios and Elkind, Edith and Wooldridge, Michael},
  journal={Synthesis Lectures on Artificial Intelligence and Machine Learning},
  volume={5},
  number={6},
  pages={1--168},
  year={2011},
  publisher={Morgan \& Claypool Publishers}
}

@book{osborne1994course,
  title={A course in game theory},
  author={Osborne, Martin J and Rubinstein, Ariel},
  year={1994},
  publisher={MIT press}
}

@article{shapley1953value,
  title={A value for n-person games},
  author={Shapley, Lloyd S and others},
  year={1953},
  publisher={Princeton University Press Princeton}
}

@article{banzhaf1964weighted,
  title={Weighted voting doesn't work: A mathematical analysis},
  author={Banzhaf III, John F},
  journal={Rutgers L. Rev.},
  volume={19},
  pages={317},
  year={1964},
  publisher={HeinOnline}
}

@inproceedings{lundberg2017unified,
  title={A unified approach to interpreting model predictions},
  author={Lundberg, Scott M and Lee, Su-In},
  booktitle={Proceedings of the 31st International Conference on Neural Information Processing Systems},
  pages={4768--4777},
  year={2017}
}

@inproceedings{datta2016algorithmic,
  title={Algorithmic transparency via quantitative input influence: Theory and experiments with learning systems},
  author={Datta, Anupam and Sen, Shayak and Zick, Yair},
  booktitle={2016 IEEE symposium on security and privacy},
  pages={598--617},
  year={2016},
  organization={IEEE}
}

@inproceedings{zhang2021interpreting,
  title={Interpreting multivariate shapley interactions in dnns},
  author={Zhang, Hao and Xie, Yichen and Zheng, Longjie and Zhang, Die and Zhang, Quanshi},
  booktitle={Proceedings of the AAAI Conference on Artificial Intelligence},
  volume={35},
  number={12},
  pages={10877--10886},
  year={2021}
}

@inproceedings{li2022fine,
  title={Fine-Grained Semantically Aligned Vision-Language Pre-Training},
  author={Li, Juncheng and HE, XIN and Wei, Longhui and Qian, Long and Zhu, Linchao and Xie, Lingxi and Zhuang, Yueting and Tian, Qi and Tang, Siliang},
  booktitle={Advances in Neural Information Processing Systems},
  year={2022}
}

@inproceedings{chen2022explore,
  title={Explore Inter-contrast between Videos via Composition for Weakly Supervised Temporal Sentence Grounding},
  author={Chen, Jiaming and Luo, Weixin and Zhang, Wei and Ma, Lin},
  booktitle={Proceedings of the AAAI Conference on Artificial Intelligence},
  volume={36},
  number={1},
  pages={267--275},
  year={2022}
}

@inproceedings{huang2021cross,
  title={Cross-sentence temporal and semantic relations in video activity localisation},
  author={Huang, Jiabo and Liu, Yang and Gong, Shaogang and Jin, Hailin},
  booktitle={Proceedings of the IEEE/CVF International Conference on Computer Vision},
  pages={7199--7208},
  year={2021}
}

@inproceedings{wang2021visual,
  title={Visual co-occurrence alignment learning for weakly-supervised video moment retrieval},
  author={Wang, Zheng and Chen, Jingjing and Jiang, Yu-Gang},
  booktitle={Proceedings of the 29th ACM International Conference on Multimedia},
  pages={1459--1468},
  year={2021}
}

@article{oord2018representation,
  title={Representation learning with contrastive predictive coding},
  author={Oord, Aaron van den and Li, Yazhe and Vinyals, Oriol},
  journal={arXiv preprint arXiv:1807.03748},
  year={2018}
}


\end{document}